\documentclass[journal]{IEEEtran}
\usepackage{enumerate}
\usepackage{graphicx}
\usepackage{hyperref}
\usepackage{comment}
\usepackage{subfigure}
\usepackage{amssymb}
\usepackage[mathscr]{euscript}
\usepackage{wrapfig}
\usepackage{multicol,listings,multirow}
\usepackage{makecell}
\usepackage{tabularx,booktabs} 
\usepackage{lipsum}
\usepackage[export]{adjustbox}
\usepackage{xcolor}
\usepackage{algorithm}
\usepackage[noend]{algpseudocode}
\makeatletter
\algrenewcommand\ALG@beginalgorithmic{\small}
\def\algbackskip{\hskip-\ALG@thistlm}
\makeatother
\algnewcommand\And{\textbf{and}}
\algnewcommand\Or{\textbf{or}}
\let\oldReturn\Return
\renewcommand{\Return}{\State\oldReturn}

\ifCLASSINFOpdf
\else
\fi
\begin{document}
%
%
\title{Incorporating Domain Knowledge To Improve Topic Segmentation Of Long MOOC Lecture Videos}
%
%

\author{Ananda Das and~Partha Pratim Das
}

%
%

\markboth{Journal of \LaTeX\ Class Files,~Vol.~14, No.~8, August~2015}%
{Shell \MakeLowercase{\textit{et al.}}: Bare Demo of IEEEtran.cls for IEEE Journals}
%



\maketitle

\begin{abstract}
Topical Segmentation poses a great role in reducing search space of the topics taught in a lecture video specially when the video metadata lacks topic wise segmentation information. This segmentation information eases user efforts of searching, locating and browsing a topic inside a lecture video. 
In this work we propose an algorithm, that combines  state-of-the art language model and domain knowledge graph  for automatically detecting  different coherent topics present inside a long lecture video. We use the language model on speech-to-text transcription to capture the implicit meaning of the whole video while the knowledge graph provides us the domain specific dependencies between different concepts of that subjects. Also leveraging the domain knowledge we can capture the way instructor binds and connects different concepts while teaching, which helps us in achieving better segmentation accuracy. We tested our approach on NPTEL\cite{nptel} lecture videos and holistic evaluation shows that it out performs the other methods described in the literature.
\end{abstract}


%
\IEEEpeerreviewmaketitle

\section{Introduction}
%
%
%
%
Online learning specially Massive Open Online Learning (MOOCs) is becoming popular as it is designed to facilitate large number of participants by providing free online access of the high quality educational contents\cite{kaplan2016higher}. So many MOOCs provider like NPTEL\cite{nptel} offers long lecture videos having duration $\sim$1 hour. These videos contain multiple non overlapping coherent topics as they are taught by the instructor. Having an index built  using these topics help user in browse, locate and searching a desired topic inside a lecture video as well as inside the whole lecture series. 
Usually these video metadata lacks the presence of annotated metadata, which supposed to hold the time stamp of each topics taught in these video. So for a learner, to search, locate and browse a topic inside the lecture series, (s)he needs to go through the whole lecture video, even in the worst case scenario s(he) requires to view all the video in the lecture series. Successful construction of an indexing system based on all the topics taught inside a lecture video requires to have topic name and its boundary information inside a lecture video. In order to facilitate users with a faster content search facility inside the whole course as well as in a video and for providing better learning experience, lecture video needs to be topic wise segmented and the video metatdata should have topic boundary information in the form of starting and ending time. Later an indexing framework can be easily built using these segmented topics which in turn mitigate user effort of topic wise video content search.

In literature several researchers \cite{lin2004segmentation,mikolov2013efficient,shah2015trace,repp2008browsing,chen2014multi} performed lecture video segmentation by measuring the structural similarity of the resources. For the text data the idea primarily revolves around dividing the whole text into fixed length overlapping text blocks and measuring the similarity between all pair of adjacent text blocks.  Shah \textit{et al.} \cite{shah2014atlas} represented cue phrases using n-gram based language model and Galanopoulos \textit{et al.} \cite{galanopoulos2019temporal} used word2vec based word embedding to represent text blocks. But finally both of them applied the traditional similarity measure based approach in obtaining topic boundaries. 

While on the other hand segmentation approach using video file \cite{yang2011automatic,che2013lecture,shah2014atlas,jeong2012accurate} primarily involves extracting text information from lecture slides shown inside the video and finding similarity measure in local context. These approach has following short comings.
\begin{itemize}
 \item Similarity measure based approach require lots of fine tuning for the videos having topics with different time duration.
 \item Topic segmentation that uses language model may capture the the underlined meaning of the transcribed file. But it cannot captures the global semantics of the transcript file.
 \item Domain knowledge in not used during the segmentation task hence knowledge semantics is not captured for finding the topic boundaries. 
\end{itemize}

To overcome these shortcomings, in this work we initially perform the topical segmentation using a language model. A language model tries to learn the structure of the text corpus through hierarchical representation thus learns to understand the both low-level syntactic feature and high level semantic features \cite{goldberg2017neural}. Though language model has some limitations of capturing the context of capturing underlined semantics of the whole text \cite{goodfellow2016deep}, topical segmentation using language model only looks into the local context of the data and unable to capture the global context resulting in an inaccurate boundary detection for the larger sized topics. To overcome this problem we additionally use a domain knowledge graph at the time of segmentation. knowledge graph contains different concepts, interconnections and relationship information among the concepts of a subjects. We leverage these knowledge present inside a  knowledge graph and use it during segmentation to capture the global context of the text corpus obtained from the speech-to-text transcript file. The major advantage of using knowledge graph during segmentation is the capability of capturing the global context that can detect the topic boundaries for larger topics in more accurate manner.

So in this work we have three major contributions.

\begin{itemize}
 \item During segmentation we use state-of-the art language model that is able to understand the implicit meaning of the text data and capture the semantics in local context which helps us in detecting segment boundaries of small sized topics.
 \item To capture the semantics in global context we use domain knowledge graph and leverage the knowledge present it during topical segmentation that makes possible to detect segment boundaries for larger length topics.
 \item Using the dependencies and relationship information available in a knowledge graph, we analyze the way instructor connects and binds different concepts to explain a topic, which helps us in getting more accurate topic boundary information.
\end{itemize}

\section{Literature Survey}

In previous years, several research works are proposed to deal with lecture video related issues like indexing, retrieval, segmentation and annotation. To accomplish these tasks mainly textual and visual information are used. Text information are available in form of subtitle file and speech-to-text transcript . Visual information is primarily obtained from lecture video and supplementary lecture material like lecture slides . Depending on the semantic resources used, we can divide the prior work into three category. Using only textual information, visual information or using both of them.

\subsection{Textual Information Based Segmentation}

Lin \textit{et al} \cite{lin2004segmentation} proposed a modified version of TextTiling \cite{hearst1997texttiling} algorithm to obtain segment boundaries from audio transcripts. Their idea is to divide the whole text transcript into fixed sized overlapping blocks. Similarity between two adjacent blocks are computed using noun phrases, verb class, pronouns, cue phrase as feature vectors. Segment boundary is considered if for a block pair, similarity score is  below some experimental threshold value. Shah \textit{et al} \cite{shah2014atlas} adopted similar approach to predict topic transition in textual content. They used cue phrases which are represented by N-gram based model while Galanopoulos \textit{et al} \cite{galanopoulos2019temporal} used word2Vec \cite{mikolov2013efficient} based word embedding to measure semantic similarity between two adjacent text blocks. In an another work Shah \textit{et al} \cite{shah2015trace} modified the above mentioned approach with the help of wikipedia text. They computed similarity between a wikipedia topic and  each fixed size text block obtained from sub-title file. Based on all similarity score and threshold value segment boundary was decided. In their work Repp \textit{et al} \cite{repp2008browsing} proposed a lecture video indexing framework by forming chain index of keywords. They extracted  a set of keywords from the speech transcripts.  For each keyword, using its occurrence position a keyword chain is formed. Finally time stamp of each chain is used to index the lecture video. Using Multi Modal Language Model Chen \textit{et al} \cite{chen2014multi} focused  to index textual data obtained from slides and speech in lecture videos, and subsequently employed a multi-modal probabilistic ranking function for lecture video retrieval. Basu \textit{et al} \cite{basu2016videopedia} built an Lecture video recommendation system for educational blogs using topic modeling technique.

\subsection{Visual Information Based Segmentation}

Researchers worked with visual contents, primarily focused on extracting textual information present in lecture slides and applying NLP based methods on them to find topic segment boundaries.  Che \textit{et al.} \cite{che2013lecture} and Baidya \textit{et al} \cite{baidya2014lecturekhoj} analyzed textual contents available in the lecture slides shown in the lecture video. SWT was used to obtain text lines form slide frame and OCR is used to recognize text data from these text lines. NLP based methods are applied on text data to perform topic segmentation.  Yang \textit{et al} \cite{yang2011automatic} used similar approach as above but used DCT based coefficient to identify text lines in a slide frame.  Che \textit{et. al} \cite{che2013lecture} segmented lecture video with synchronized slides. By analyzing it's OCRed text they partition the slides into different subtopics by examining their logical relevance. Slides are synchronized with the video stream, the subtopics of the slides indicate exactly the segments of the video. Another group of researches tried to match  between video frames and lecture slides, assuming that lecture slides are available as supplementary material.  Eberts \textit{et al.} \cite{eberts2015amigo} matched between lecture slides and video frames to create localization information of each slide shown in the lecture video. They leverage alignment of the presentation with the reading order of its  supplementary material. They handled it by using a combination of  SIFT local feature and temporal model. As temporal model they used Hidden Markov Model and heuristic filters.   Ma \textit{et al.} \cite{ma2017lecture} Analyzed lecture slides to get topic segments. These segment boundaries are mapped to the lecture video by matching lecture slides and video frames. They handled different perspective views, resolution, illumination, noise, and occlusions in videos by using boosted margin maximizing neural networks.  In some work, a lecture video is considered as a sequence of events like displaying lecture slides or the instructor. Assuming change of event as the change of current topic, topic boundary is determined by detecting the event change.  Shah \textit{et al.} \cite{shah2014atlas}  detected event change in lecture videos and considered change of an event is change of a topic. They consider events as displaying instructor or displaying lecture slide in the video. To detect event change they trained an SVM by computing color histogram of video frames.  Jeong \textit{et al.} \cite{jeong2012accurate} handled lecture video where recording was done by a non stationary camera settings. They identified topic change by computing frame similarity between two adjacent frames. based on some threshold value topic boundary was identified. Ma \textit{et al} \cite{ma2012lecture} used color histogram of video frames to identify slide transition, which in turn is used to detect topic change.

\subsection{Joint Approach of Segmentation}

To obtain more accurate result researchers adopted fusion based method, which involves merging two sets of segmentation information obtained from heterogeneous sources like speech transcripts, video content, subtitle file or OCRed text. Basic idea behind the fusion approach is replacing two segment boundaries by their average transition time if two transition occurs within thirty seconds. Shah \textit{et al.} \cite{shah2014atlas, shah2015trace} used both subtitle file and video frames and wikipwdia text to get final segmentation. In \cite{yang2014content} content based lecture video retrieval system is proposed by extracting keywords from slide text line and analyzing OCR and ASR transcripts of the video.

\section{Methodology}

\begin{figure}[!h]
\centering
\includegraphics[width=0.48\textwidth]{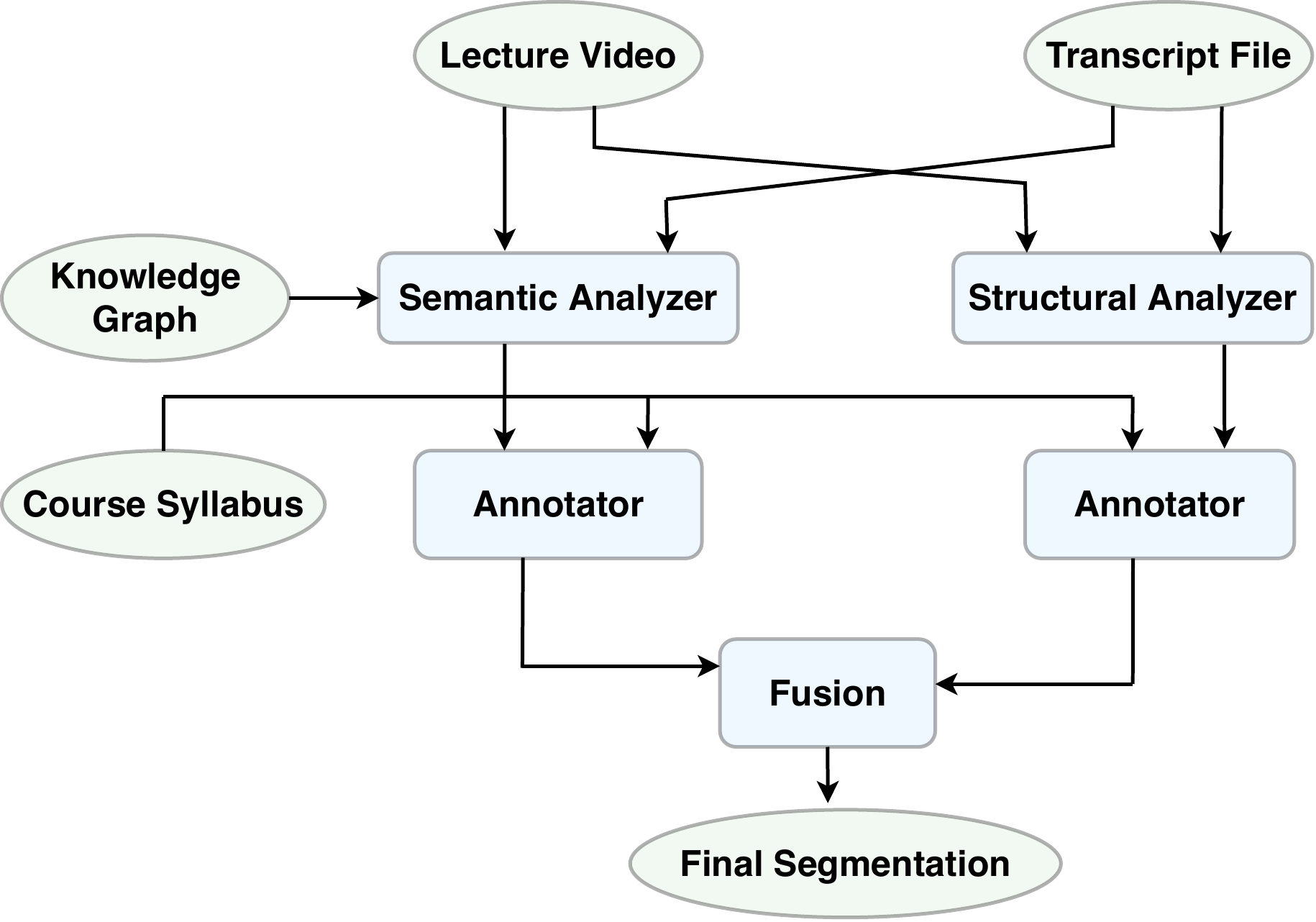}
\caption{Schematic diagram of the methodology}
\label{fig:method}
\end{figure}

Different steps to perform topical segmentation of long lecture videos are mentioned below and the schematic diagram is shown in the Figure~\ref{fig:method}.

\begin{enumerate}
 \item \textbf{Segmentation using structural analysis} 
In this method of lecture video segmentation first we segment the transcript file associated with a lecture video. During segmenting the transcript file we analyze the structure of the textual information and try to understand the implicit meaning of the whole transcript file. For understanding the transcribed language we train a language model and reuse it during the actual segmentation work. After we obtain the segmentation in transcript file domain we reversely map it to the lecture video domain. Section~\ref{sec:structural} describes detail steps of this task.
 
 \item \textbf{Segmentation using Semantic Analysis} In this step we try to obtain topic boundary information by leveraging the knowledge semantics present inside a domain knowledge graph. Also we capture the underlined meaning of the transcript file by using a state-of-the art language model. Section~\ref{sec:semantic} describes step by step procedure to perform semantic analyzing for obtaining the topic boundaries.
 
 \item \textbf{Annotation} To distinguish a segmented topic from others and to make them searchable we need to asisgn a name to each of the segmented video. We call this step as segment annotation. we create annotation module whose responsibility is to determine topic name with the help of course syllabus and lecture videos. Details annotation process are described in Section~\ref{sec:annotation}. 
 
 \item \textbf{Fusion} 
 As mentioned earlier we get two set of topic boundary information. One using by structural analysis of the transcript file and another set of boundary information is obtained from semantic analysis based segmentation which essentially leverages domain knowledge from a knowledge graph. In this step we fuse these two set of boundary information to obtain better segmentation results. The fusion step is described in details in Section~\ref{sec:fusion}

\end{enumerate}

\section{Segmentation Using  Structural Analysis} \label{sec:structural}

The basic idea behind the structural segmentation is to identify few sentences those can be considered as topic boundaries. For doing this, we can consider a transcript file as a sequence of sentences, where each sentence consists of fixed number of words. Now semantic segmentation of text can be formulated as a binary classification problem. Given a sentence $ {S_{K}} $ with \textit{K} sentences $ {S_{0}, S_{1}, .... , S_{K-1}} $ preceding it and \textit{K} sentences $ {S_{K+1}, S_{K+2}, .... , S_{2K}} $ following it, we need to classify whether sentence $ {S_{K}} $ is the beginning of a segment or not. To represent each sentence in vector form we use InferSent \cite{conneau2017supervised} approach. Next, each encoded sentence along with its K-sized left and right contexts are fed to a classifier, whose output helps us to determine segment boundaries of each topic. After segment boundaries are detected in transcript file we use them for finding the topic boundary in the lecture video. We perform the following steps to accomplish the segmentation task.

\subsection{Sentence Encoding}

\subsubsection{Sentence Encoding}
We first divide the whole transcript file into strings of fixed length ($20$ words). To create sentence encoding we have used pre-trained sentence encoder model proposed by InferSent \cite{conneau2017supervised}. It tries to build a universal sentence encoding by learning the sentence semantics of Stanford Natural Language Inference (SNLI) \cite{bowman2015large} corpus. SNLI corpus contains $ 570K $ human generated English sentence pairs. Each sentence pair is manually labeled and belongs to one of the three categories: entailment, contradiction and neutral. These corpus are used to train a classifier on top of a sentence encoder, which encodes both the sentences in a sentence pair in similar manner. Semantic relationship between these two sentence vectors are computed which is turn passed through a dense layer and soft-max layer. Finally the model classifies the sentence pair among one of the three mentioned categories. Conneau et al. \cite{conneau2017supervised} adopted a bi-directional LSTM with a max-pooling operator as sentence encoder. In our task we use this pre-trained sentence encoder model to create sentence vectors, which are used in later stages.

\subsection{Sentence Classification} \label{subsec:sentClass}

At this stage we decide, given a sentence with its $K$ sized left and right contexts, whether the sentence is the beginning of a segment or not. Our classification approach is similar in spirit of the method proposed by Badjatiya et al. \cite{badjatiya2018attention} with some modifications. Relationship of a sentence with its left and right contexts, plays most important role to become the sentence as the segment boundary. Hence, first we compute context vectors of left and right contexts using stacked bi-LSTM with an attention layer. In stacked bi-LSTM multiple layers of both forward and backward pass LSTM \cite{hochreiter1997long} are stacked upon together to obtain better accuracy. Also two sentences located at equal distance, one in the left context and other on the right, might not have similar impact on the middle sentence to be it the segment boundary. To handle this we apply a soft attention layer on top of the bi-LSTM layer. Attention allows to assign different weights to the different sentences resulting better model accuracy. We define attention vector, $ Z_{s} $ as 

$ e_{i}^{K\times 1} = H_{i}^{K\times d} \times W^{d\times 1} + b_{i}^{K\times 1} $ , $ a_{i} = exp(tanh(e_{i}^{T}Z_{s})) $, $ \alpha_{i} = \frac{a_{i}}{\sum_{p} a_{p}} $, $ v_{i} = \sum_{j=1}^{K} \alpha_{j}h_{j} $ 

where, $d$ is the dimension of the output vector of last bi-LSTM layer, $H$ is the output of the bi-LSTM layer, $W$ and $b$ are the weight matrix and bias vector and $v_{i}$ is the resultant context embedding. After computing both the context vector in similar manner, we find semantic relatedness between left context (let it be $C_{L}$), right context ($C_{R}$) and the middle sentence ($M$) by the following way. 

\begin{enumerate}
 \item Concatenation of three representation $ ( C_{L}, M, C_{R} )$
 \item Pair wise multiplication of the vectors $( C_{L}*M, M*C_{R}, C_{R}*C_{L} )$
\end{enumerate}

\begin{figure}
\centering
\includegraphics[width=0.49\textwidth]{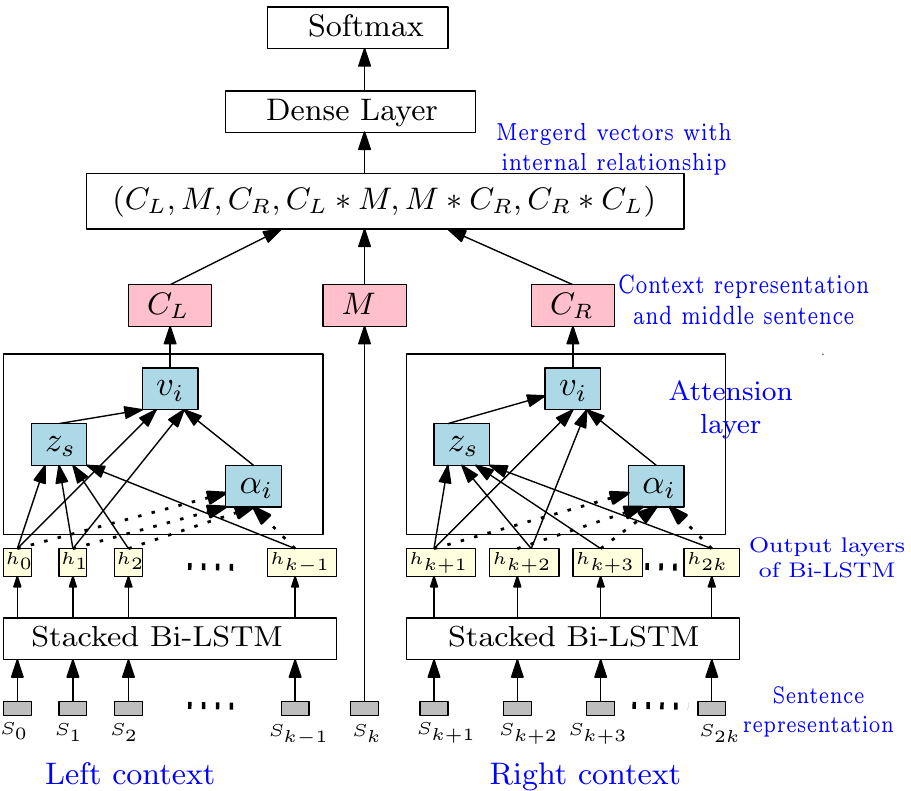}
\caption{Sentence Classifier}
\label{fig:Seg}
\end{figure}

The resultant vector which contains relationship information among these vectors, is fed to a binary classifier consists of fully connected layers culminating in a soft-max layer. The architecture diagram of this approach is is shown in Figure~\ref{fig:Seg}. Sentences classified as segment boundaries are used to obtain topic boundary in next step.

\subsection{Text To Video Mapping} 
\begin{figure}[!h]
\centering
 \fbox{\includegraphics[width=0.48\textwidth]{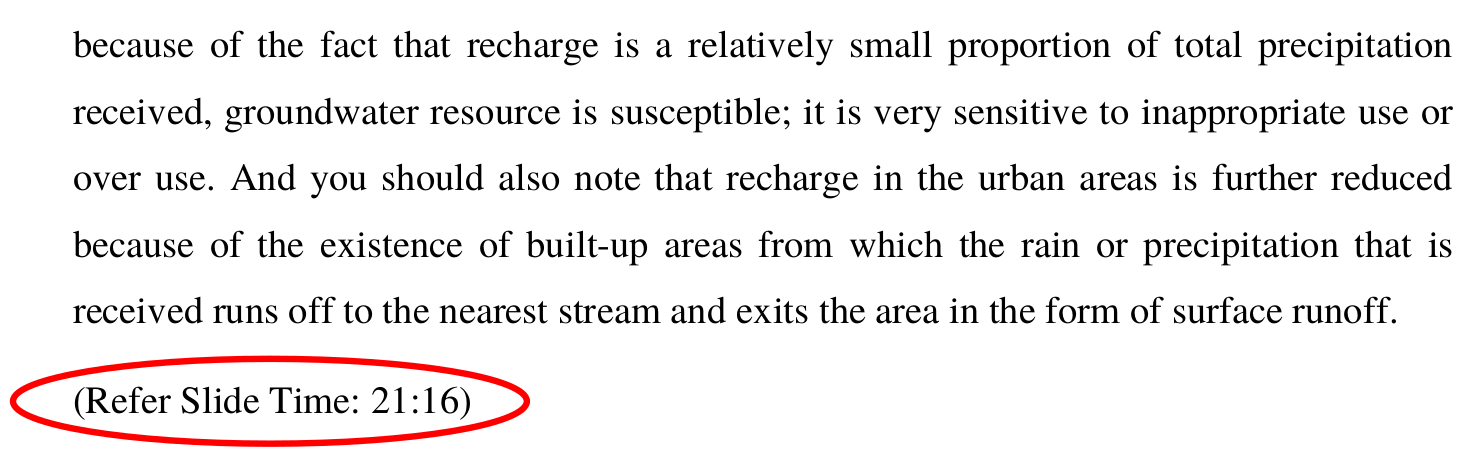}}

 \caption{Mapping Information Sample: Time stamp in shown in the red box}

 \label{fig:mappingExample}
\end{figure}

In this step, having segment boundaries obtained from text transcript, we need to identify corresponding time stamp in the original lecture video. These identified time stamps will act as starting time and ending time of different topics. We take help from text to video mapping information present inside speech-to-text transcript. Mapping information is available in the following form. After some random  sentence it contains time stamp information when it is pronounced by the instructor in the lecture video. A sample of such file is shown in \ref{fig:mappingExample}. We perform text to video mapping in the following way. Consider two consecutive time stamps $t_{S}$ and $t_{S+1}$ are given in the text transcript. Let, there exists $W$ words in between these two time stamps. Say, we get a segment boundary (in form of a sentence) inside this window. If there are $W_{p}$ words preceding this sentence, then video time stamp corresponding to this sentence is $\frac{t_{S+1} - t_{S}}{W} * W_{p}$. For a lecture video if we get $N$ ($TS_{1}$, $TS_{2}$, $TS_{3}$ ... $TS_{N}$) such time stamp, then we conclude that the video contains $N-1$ topic with starting and ending time pair as {($TS_{1}, TS_{2}$), ($TS_{2}, TS_{3}$) .... ($TS_{N-1}, TS_{N}$)}.
\begin{figure}[!h]
\centering
 \fbox{\includegraphics[trim=0 0 120 0, clip, width=0.48\textwidth]{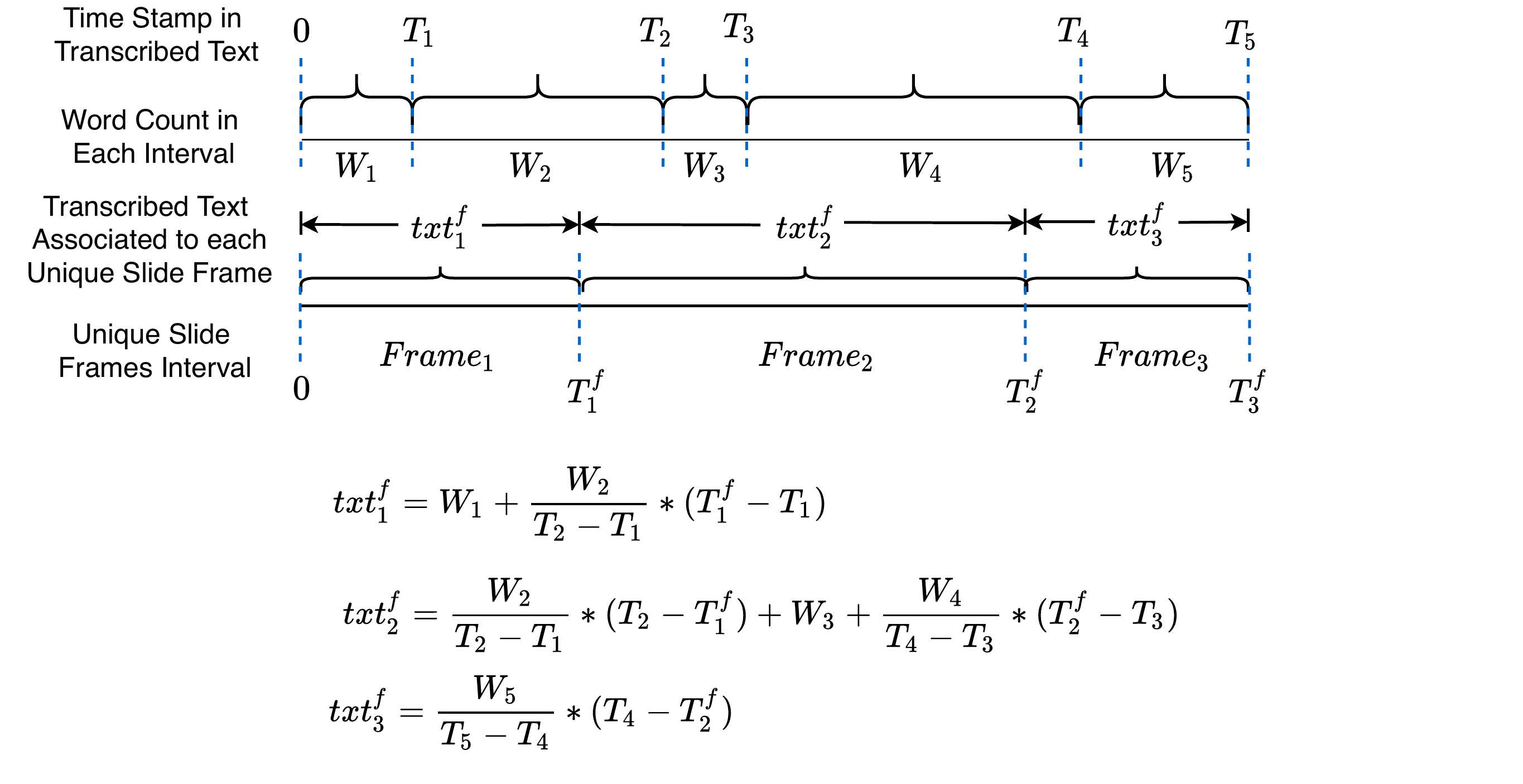}}
 \caption{Sample illustration of finding transcribed text associated to each unique slide frame.}
 \label{fig:frameText}
\end{figure}

\subsection{Model Training and Parameter Tuning}

To train sentence classifier we use Wikipedia \cite{wiki} pages. Section boundaries in the Wikipedia pages are considered as segment boundaries. We neglect the introduction section of each page, as it briefs the overall idea of the whole page. We have randomly chosen $\sim$300 Wikipedia pages having $\sim$55K sentences, with $\sim$8\% sentences being the segment boundaries. We handle this highly imbalanced dataset by selecting weighted binary cross entropy loss function, which gives higher weight to the minority class and lower weight to the majority class. This loss function $\ell$, can be defined as,
$$ \ell  = -\frac{1}{N} \Sigma_{n=1}^{N} w_{n}(t_{n}log(o_{n}) + (1-t_{n})log(1-o_{n})) $$
where, $w_{n}$: weighting factor associated with loss of a class. $t_{n} \in {0,1}$: Binary label of the target class. $o_{n} \in [0,1]$: Classification probability score produced by the model. $N$: Total number of class, here $N = 2$

We use AdaDelta \cite{zeiler2012adadelta} optimizer with dropout $0.3 $. Epoch size is chosen as $35$ and the model is trained with mini batch size as $40$.

\section{Segmentation Using Semantic Analysis} \label{sec:semantic}

\begin{figure}[!h]
\centering
 \fbox{\includegraphics[trim=62 165 69 50, clip, width=0.47\textwidth]{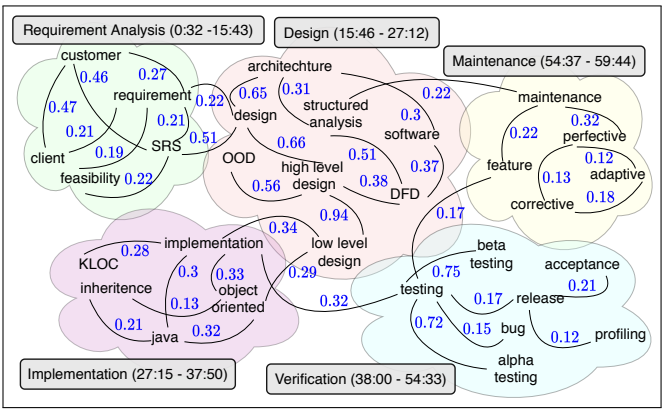}}
 \caption{Illustration of knowledge graph based topical segmentation of lecture videos. All the concepts covered in a lecture video are shown. Edge weight represents contextual semantic similarity among the concepts. Each topic, represented as a cloudy shape, consists of multiple concepts. Text in each rectangular box represents corresponding topic name and its interval.}
 \label{fig:ovearAll}
\end{figure}

THe basic idea of this approach is to perform topical segmentation of long lecture videos by leveraging the inherent knowledge available in a domain knowledge graph($\mathscr{KG}$). Also, at the time of teaching, instructor has his own way of selecting different concepts, connecting and combining them together to explain a topic. We combine these two information to perform topical segmentation of long MOOC lecture videos.  We take NPTEL lecture videos, each of them is synchronized with lecture slides and associated with speech-to-text transcript. We identify unique slides shown in the video, their time interval and associated textual information from the transcript file. Using these information we construct graphlets, each represents a unique lecture slide shown in the lecture video. In each graphlet, the nodes are the concepts used by the instructor, an edge indicates presence of a direct connection between two concepts in the  $\mathscr{KG}$ and edge weight represents the contextual semantic similarity between two adjacent concepts of that edge. Here `contextual semantic similarity' means semantic similarity between two concepts in the context of the way instructor combines them together during teaching. The basis of our work is to analyze all these graphlets and find how concept change occurs between two adjacent graphlets. Finally, measuring the amount of concept change and contextual semantic similarity among concepts, our system provides boundaries (starting-time and ending-time) of different topics taught in a lecture video. The idea is schematically shown in Figure~\ref{fig:ovearAll} where we take a lecture video on `Overview of waterfall model'. The diagram represents different concepts used by the instructor which are distributed into five different topics. Connectivities between these concepts are taken from the $\mathscr{KG}$ and edge weight represents contextual semantic similarity among concepts. After analyzing all the graphlets created from the unique slides, we can perform topical segmentation and mark the topic boundaries. 

Steps to perform segmentation using semantic analysis are described below.

\subsection{Edge Weight Computation} \label{subsec:edgeWeightcomp}

In this step, we compute the edge weight between different concepts present in the $\mathscr{KG}$. Essentially, edge weight represents how two concepts are semantically similar with each other in the context of transcript file. We find semantic similarity between them considering the context of their co-occurrence. To determine the context we use FLAIR \cite{akbik2019flair} api that runs on top of the BERT \cite{devlin2018bert} model. Unlike static word embedding, BERT can produce dynamic word embedding considering the sentences in which a word has occurred. Here in our work, we capture the contextual semantic similarity between two concepts, so that we can analyze how instructor correlates and connects different concepts during teaching. We observe, a concept might occur multiple times in the transcript file. So, during similarity score computation, we consider all pairwise co-occurrences between two given concepts. We take the text chunk that contains first such co-occurrence and send it to the FLAIR api to find $2048$ dimensional word embedding of each concept and calculate cosine similarity between them. We take all the co-occurrences between any two concepts, find multiple similarity scores between them and take the average to find final contextual similarity between them. After similarity computation is done, we use it as the edge weight in the $\mathscr{KG}$ if there exist a link between these two concepts. We maintain a dictionary which holds the contextual semantic similarity score between all pair of concepts present in the transcript file. 

\subsubsection{Slide Graph Construction} 

First we construct an un-directed weighted graph representing each unique slide frame and we call it \textit{slide\_graph}. To construct $n^{th}$ \textit{slide\_graph} ($G^{s}_{n}$) we use transcribed text (say, $TXT_{n}$) associated with $n^{th}$ unique slide, the $\mathscr{KG}$ and the edge weight dictionary created in Section~\ref{subsec:edgeWeightcomp}. Construction steps are given below. 

\begin{enumerate}
 \item A vertex is created for each concept word in $TXT_{n}$, set vertex name as the concept and vertex weight as the term frequency of that concept in $TXT_{n}$.
 \item We draw an edge between two vertices if there is a corresponding edge present in the $\mathscr{KG}$. Edge weight is assigned using the dictionary created in Section~\ref{subsec:edgeWeightcomp}.
 \item We remove direction information from the generated graph because during graph analysis in the next phase, a node may not be reachable from the other nodes, though there exists a strong connection among them. 
 \item If the generated graph is not connected, we draw an edge between two connected components whose weight is the maximum among all the possible edges between these two components.
\end{enumerate}

\subsection{Finding Potential Topic Boundary}

In this step, we determine potential topic boundaries by analyzing all the \textit{slide\_graph}. During lecture, instructor changes current slide at the time of going to a different topic, but vice versa is not necessarily true. In fact usually multiple slides are used for explaining a particular topic and few consecutive slides are closely related. Here closely related means, these consecutive slides represent almost same concepts. In this step we identify closely related consecutive slides and merge them together. Also, we mark the slides which are not close and those slide changes might be the topic boundaries. We measure how much concept changes have been occurred between two consecutive lecture slides. We define \textit{concept\_change\_score}, $CCS(j,j+1)$ between two consecutive \textit{slide\_graph},  $G^{s}_{j} = (V^{s}_{j},E^{s}_{j})$ and $G^{s}_{j+1} = (V^{s}_{j+1},E^{s}_{j+1})$ as follows.

\[ CCS(j,j+1) = \left\{
    \begin{array}{lr}
        0  ~~~~ : \{V^{s}_{j} \cap V^{s}_{j+1}\} = \phi\\
        \frac{total~minimum~term~frequency}{total~maximum~term~frequency} ~~~~ : Else
    \end{array}
    \right.
\]
where, 
$total~minimum~term~frequency = \sum\limits_{k=0}^{N}min(tf(v_{k} \mid v_{k} \in V^{s}_{j}),tf(v_{k} \mid v_{k} \in V^{s}_{j+1}))$

\begin{figure}[!h]
\centering
\subfigure[]{

\includegraphics[trim=0 299 150 132, clip,width=0.49\textwidth]{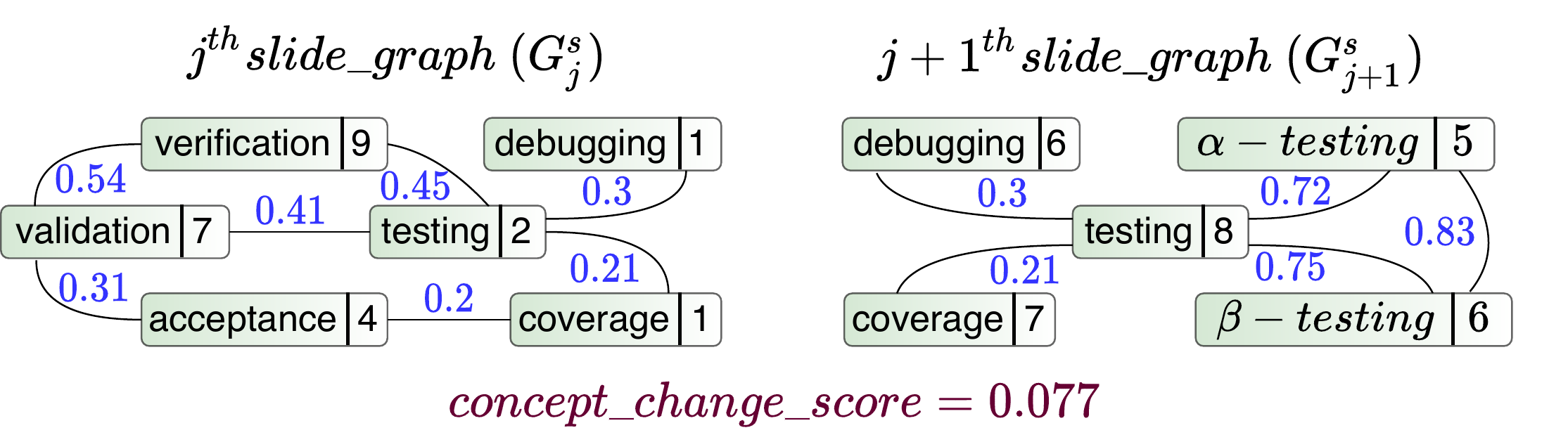} \label{fig:slideTrans-1}
}
\subfigure[]{
\includegraphics[trim=45 239 140 212, clip,width=0.49\textwidth]{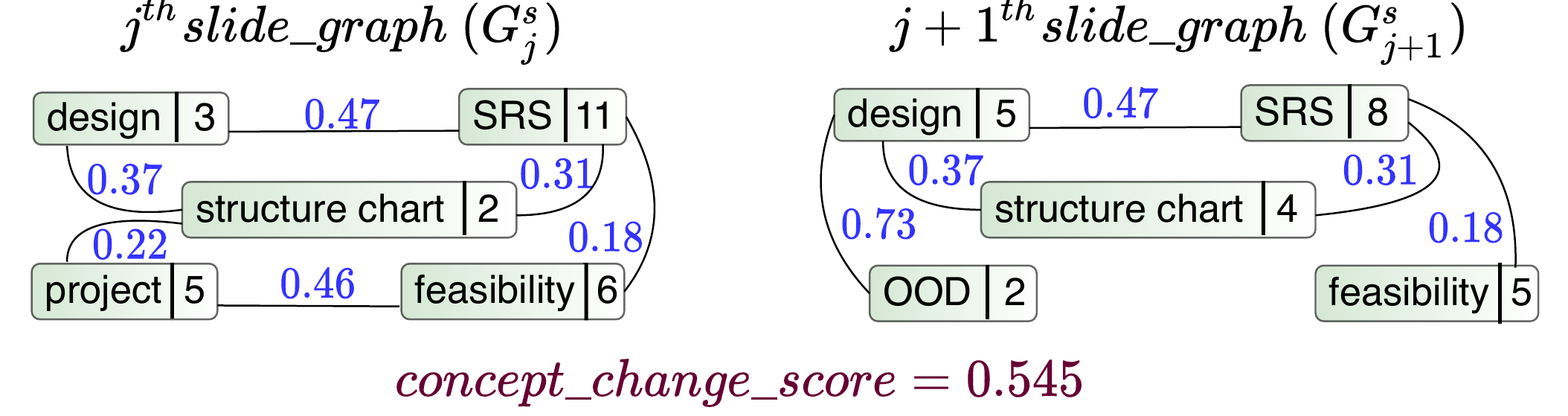} \label{fig:slideTrans-2}
}
\caption{An example of \textit{concept\_change\_score} between two consecutive \textit{slide\_graph}. Each node represents a concept and its term frequency, edge weights indicate contextual semantic similarity among concepts. Slide changes shown in Figure~\ref{fig:slideTrans-1} is a potential candidate of being a topic boundary as most of the concepts are different and $CCS(j,j+1)$ is low ($0.077$). While, Figure~\ref{fig:slideTrans-2} shows $G^{s}_{j}$ and $G^{s}_{j+1}$ are closely related as most of the concepts present in these two slides are same and $CCS(j,j+1)$ is large ($0.545$) compared to the other.} \label{fig:CCS}
\end{figure}

$total~maximum~term~frequency=\sum\limits_{k=0}^{M}max(tf(v_{k} \mid v_{k} \in V^{s}_{j}),tf(v_{k} \mid v_{k} \in V^{s}_{j+1}))$

where, $N = |\{V^{s}_{j} \cap V^{s}_{j+1}\}| $ and $M = |\{V^{s}_{j} \cup V^{s}_{j+1}\}| $ and $tf(v_{k} | v_{k} \in V^{s}_{j})$ indicates term frequency of vertex $v_{k}$ when $v_{k} \in V^{s}_{j}$. Here consideration of term frequency is important. We observe that, instructor puts more emphasis on few concepts while other concepts are used for giving example or reference purpose. Throughout this paper we call them as primary and auxiliary concepts respectively. Usually, primary concepts have higher term frequency than the auxiliary concepts. Also, Concepts with higher term frequency indicates that instructor has spent more time on them than the concepts with lower term frequency. Leveraging this fact we consider term frequency in \textit{concept\_change\_score} computation. Clearly \textit{concept\_change\_score} is in range between $[0,1]$ and its value is $1$ if two consecutive \textit{slide\_graph} are exactly similar and $0$ if their vertices are disjoint. An illustrative example of \textit{concept\_change\_score} is shown in Figure~\ref{fig:CCS}.

\begin{figure}[!h]
\centering
 \includegraphics[trim=5 235 140 192, clip,page=2,width=0.49\textwidth]{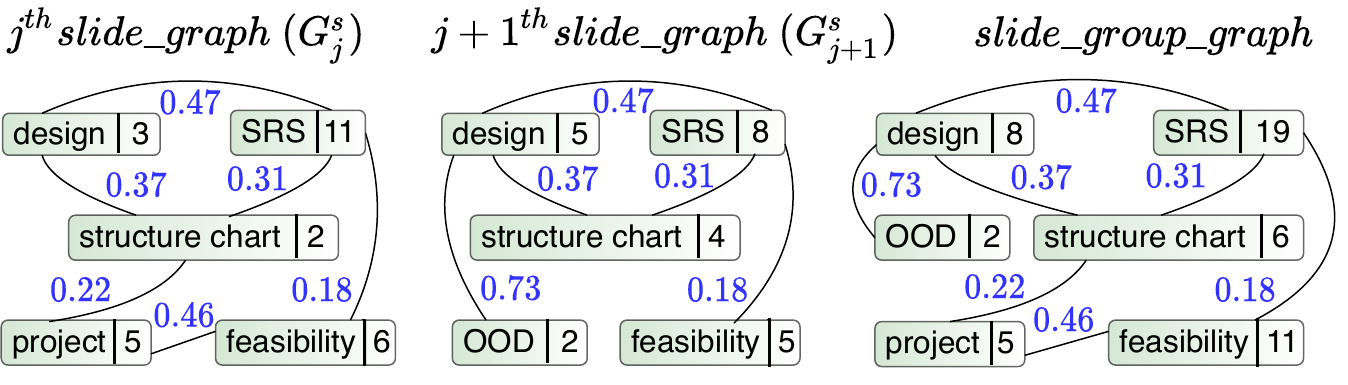}
 \caption{Merging two consecutive \textit{slide\_graph} to construct a \textit{slide\_group\_graph}. }
 \label{fig:graphMerging}
\end{figure}

We use this \textit{concept\_change\_score} to identify consecutive \textit{slide\_graph} where significant amount of concept change happens and mark those slide transitions as potential topic boundaries. We also identify closely related consecutive \textit{slide\_graph} and merge them together as those corresponding slides represent almost same concepts. Steps to perform these operations are,
 \begin{enumerate}
   \item We compute \textit{concept\_change\_score} between all pair of consecutive \textit{slide\_graph} and select those changes where the \textit{concept\_change\_score} is less than the average value. We mark these changes as potential topic boundaries.
   \item We merge those pair of consecutive \textit{slide\_graph} where \textit{concept\_change\_score} is higher than the average value. We call this merged graph as \textit{slide\_group\_graph}. A sample illustration of constructing \textit{slide\_group\_graph} is shown in Figure~\ref{fig:graphMerging}.
   \item \textit{slide\_graph} merging policy is as follows. Consider two consecutive \textit{slide\_graph}, $G^{s}_{j} = (V^{s}_{j},E^{s}_{j})$ and $G^{s}_{j+1} = (V^{s}_{j+1},E^{s}_{j+1})$ which are merged to get \textit{slide\_group\_graph}, denoted as, $G^{g} = (V^{g},E^{g})$. where, 
$V^{g} = V^{s}_{j} \cup V^{s}_{j+1}$ and $E^{g} = E^{s}_{j} \cup E^{s}_{j+1}$. Vertex weight is modified as the summation of its weight in $G^{s}_{j}$ and $G^{s}_{j+1}$.
  \end{enumerate}
 
\subsubsection{Finding Actual Topic Boundary} \label{subSec:actual}
In this step we have a collection of \textit{slide\_graph} and \textit{slide\_group\_graph}. We call each of them as \textit{cluster} where each of them represents some concepts and transition from one 
\textit{cluster} to the next one makes substantial amount of concept change. Some of these transitions are actual topic boundaries while others are not. To find actual topic boundaries, we analyze how concepts are closely connected with each other within a topic than the concepts belonging in two different topics. 

\textbf{Primary and auxiliary concepts} In our dataset we observe, in each \textit{cluster}, instructor primarily focuses on some concepts and rest are used for explanation or reference purpose. We call them as primary concepts and auxiliary concepts respectively. For example, to explain a topic, say ``Agile Methods'', instructor primarily focuses on concepts like, ``sprint'', ``scrum'', ``eXtreme Programming'', ``face-to-face conversation'' etc. But sometimes he also mentions ``waterfall'', ``spiral'' for reference purpose or ``Zoho Sprints'', ``Kanbanize'' to mention the tools used for agile. We also find that term frequency of primary concepts are higher than the auxiliary concepts. Leveraging this fact, we take only the primary concepts and their interconnection for finding actual topic boundaries. we create \textit{cluster\_centroid} from each \textit{cluster} which represents the primary concepts and the interconnections among them. In this computation we consider top 70\% concepts as primary concepts. We discuss the selection of this value in Section~\ref{subsec:threshold} . Steps to construct a \textit{cluster\_centroid} are given below.
\begin{enumerate}
    \item Sort the concepts in non increasing order according to their term frequency.
    \item In case of tie in term frequency, we consider total weight of the incident edges to a concept. Here higher weight signifies a concept is semantically closer to other concepts than that of the concept with lower weight in incident edges.
    \item Consider top $70\%$ concepts and their interconnecting edges during formation of the \textit{cluster\_centroid}. 
\end{enumerate}

To detect actual topic boundaries, we consider three consecutive (say, $i^{th}$, $i+1^{th},$ and $i+2^{th}$) \textit{cluster\_centroid} denoted as $G^{c}_{i} = (V^{c}_{i}, E^{c}_{i})$ , $G^{c}_{i+1} = (V^{c}_{i+1}, E^{c}_{i+1})$ and $G^{c}_{i+2} = (V^{c}_{i+2}, E^{c}_{i+2})$ respectively. We represent the transition from $G^{c}_{i}$ to $G^{c}_{i+1}$ as ${\tau(i,i+1)}$ and from $G^{c}_{i+1}$ to $G^{c}_{i+2}$ as ${\tau(i+1,i+2)}$. We may encounter three different cases as described below. An illustrative diagrams are also shown in Figure~\ref{fig:caseA}, Figure~\ref{fig:caseB}, and Figure~\ref{fig:caseC}. These figures contains small dots that represent different concepts and lines connecting them are the edges. Cloudy, rectangular and oval shapes represent three consecutive \textit{cluster\_centroid} $G^{c}_{i}$, $G^{c}_{i+1}$ and $G^{c}_{i+2}$ respectively. Thick gray arrows represent the transition from one \textit{cluster\_centroid} to the next one.

\begin{enumerate}
\item \textbf{Case A:} If we find both $\{V^{c}_{i} \cap V^{c}_{i+1}\} = \phi$ and $\{V^{c}_{i+1} \cap V^{c}_{i+2}\} = \phi$, we mark ${\tau(i,i+1)}$ as a topic boundary. This situation is represented in the Figure~\ref{fig:caseA}. 

\begin{figure}[!h]
\centering
    \includegraphics[width=0.41\textwidth]{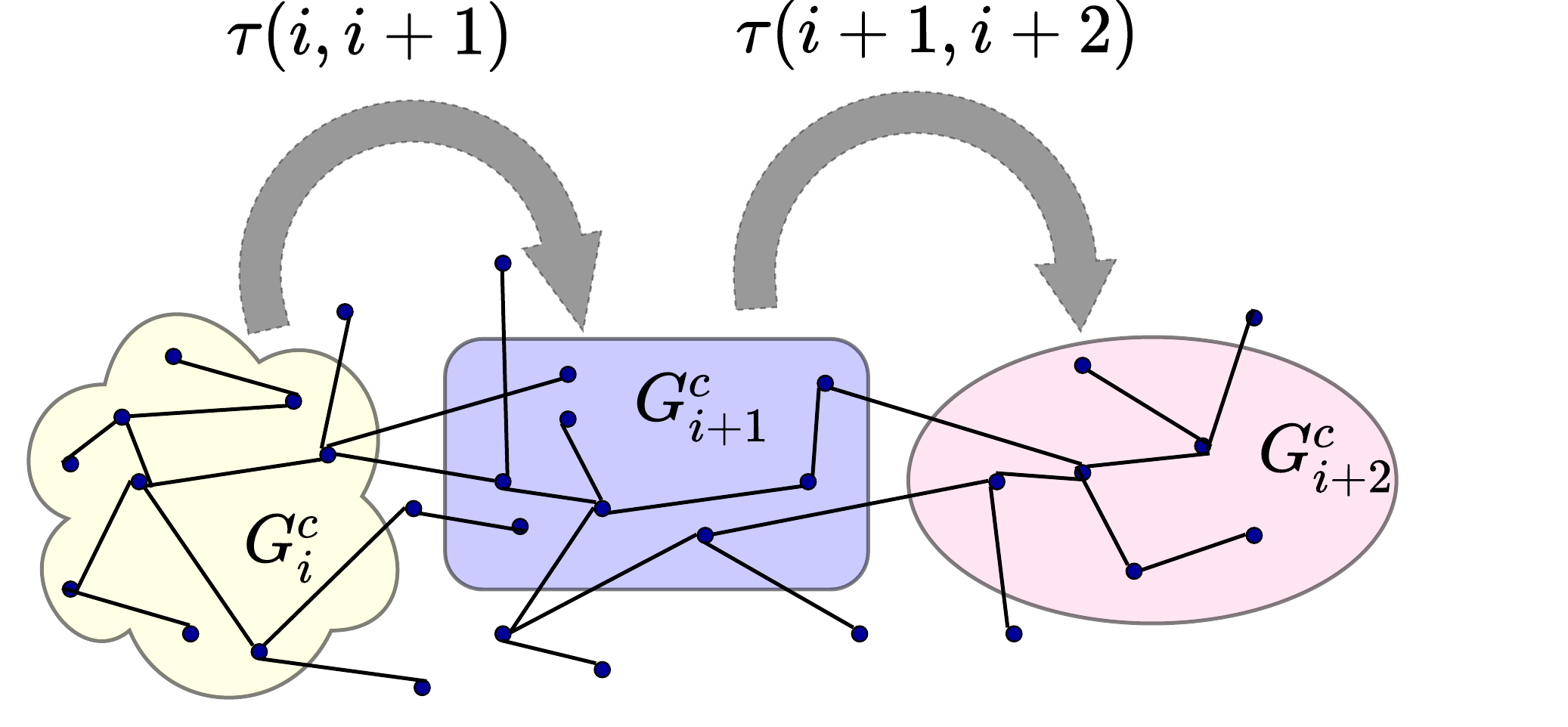}
\caption{An illustration of a scenario (Case A), occurs during teaching.}
\label{fig:caseA}
\end{figure}

\item \textbf{Case B:} $G^{c}_{i}$ and $G^{c}_{i+2}$ are in two different topics (say, $topic_{x}$ and $topic_{y}$ in chronological order). Instructor gradually changes from $topic_{x}$ to $topic_{y}$ through $G^{c}_{i+1}$. Different situations may occur like,
\begin{enumerate}
 \item \underline{Situation-1} Most of the concepts covered in $G^{c}_{i+1}$ are part of $topic_{x}$, hence $\tau(i+1,i+2)$ is a topic boundary as shown in Figure~\ref{subFig:b1}.
 \item \underline{Situation-2} Most of the concepts covered in $G^{c}_{i+1}$ are part of $topic_{y}$, hence $\tau(i,i+1)$ is a topic boundary as shown in Figure~\ref{subFig:b2}.
\end{enumerate}

\begin{figure}[!ht] 
\centering
\subfigure[Case B: Situation-1]{
{\includegraphics[trim=80 475 143 38, clip,width=0.23\textwidth]{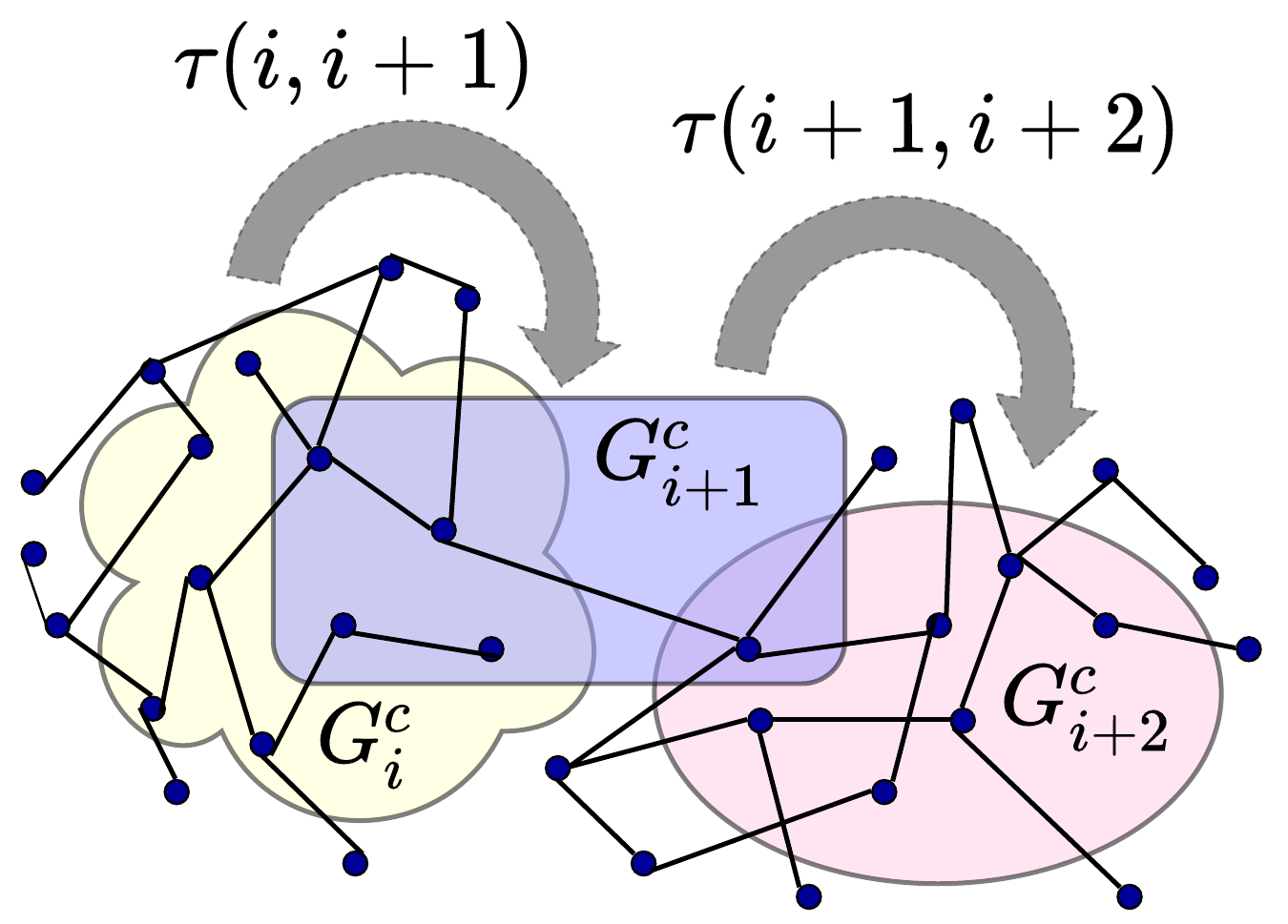}} \label{subFig:b1}
}~~~~
\subfigure[Case B: Situation-2]{
{\includegraphics[trim=107 481 128 47, clip,width=0.23\textwidth]{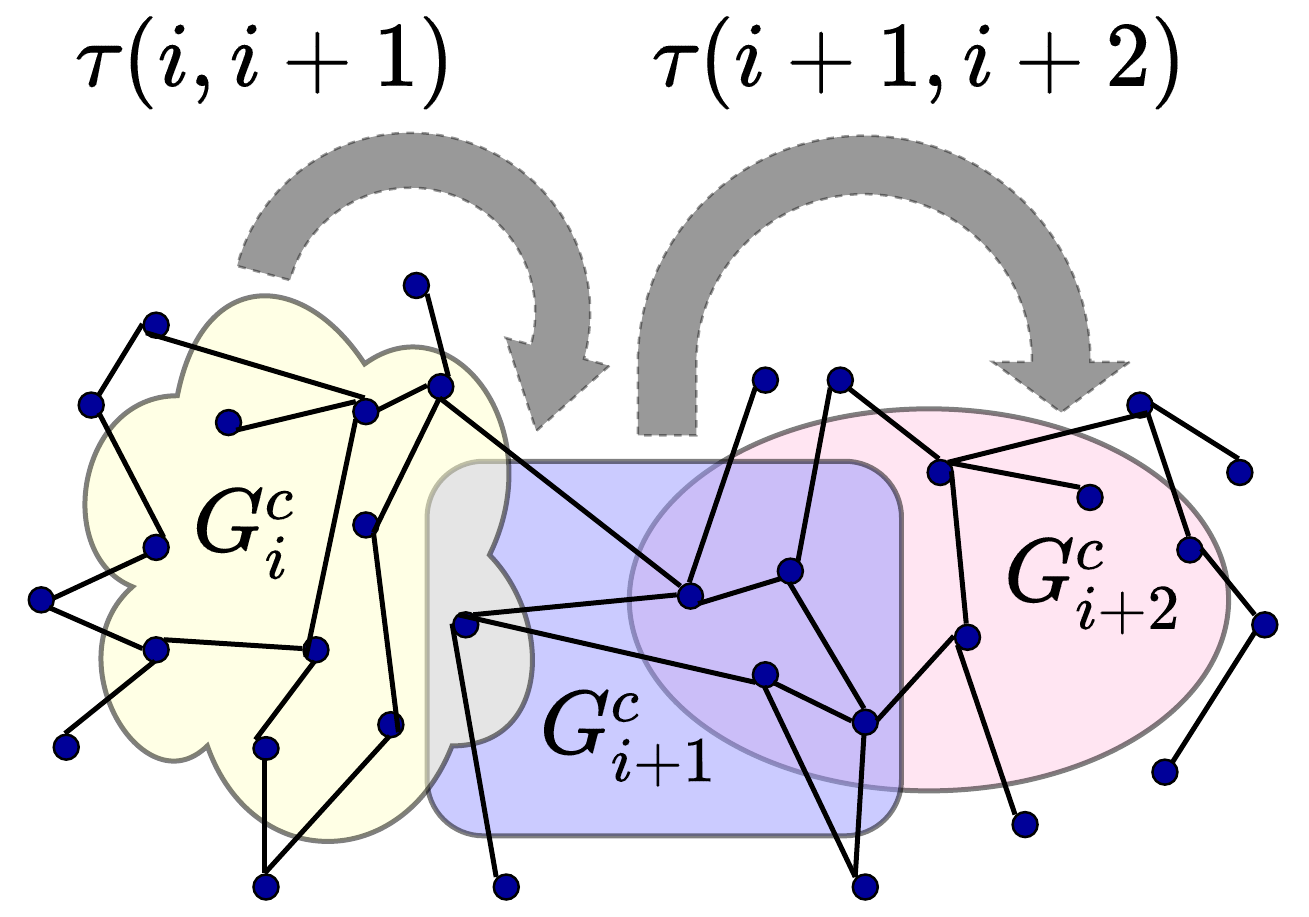}} \label{subFig:b2}
}
\caption{An illustration of the scenario (Case B), occurs during teaching.} \label{fig:caseB}
\end{figure}

\item \textbf{Case C:} $G^{c}_{i}$, $G^{c}_{i+1}$ and  $G^{c}_{i+2}$ all falls under same topic.
\begin{enumerate}
 \item \underline{Situation-1} There are some changes in concepts between $G^{c}_{i}$ and $G^{c}_{i+1}$, but in  $G^{c}_{i+2}$ instructor again back to the concepts present in $G^{c}_{i}$ as shown in Figure~\ref{subFig:c1}.
 \item \underline{Situation-2} Most of the concepts covered in a particular topic are present in $G^{c}_{i}$ and very few additional concepts are taught in $G^{c}_{i+1}$ and $G^{c}_{i+2}$ as shown in Figure~\ref{subFig:c2}.
 \item \underline{Situation-3} Similar to the Case C: Situation-2, while most of the concepts are covered in $G^{c}_{i+1}$as  shown in Figure~\ref{subFig:c3}.
 \item \underline{Situation-4} Similar to the Case C: Situation-2, while most of the concepts are covered in $G^{c}_{i+2}$ as shown in Figure~\ref{subFig:c4}.
\end{enumerate}

\begin{figure}[!ht] 
\centering
\subfigure[Case C: Situation-1]{
{\includegraphics[trim=150 407 172 107, clip,width=0.20\textwidth]{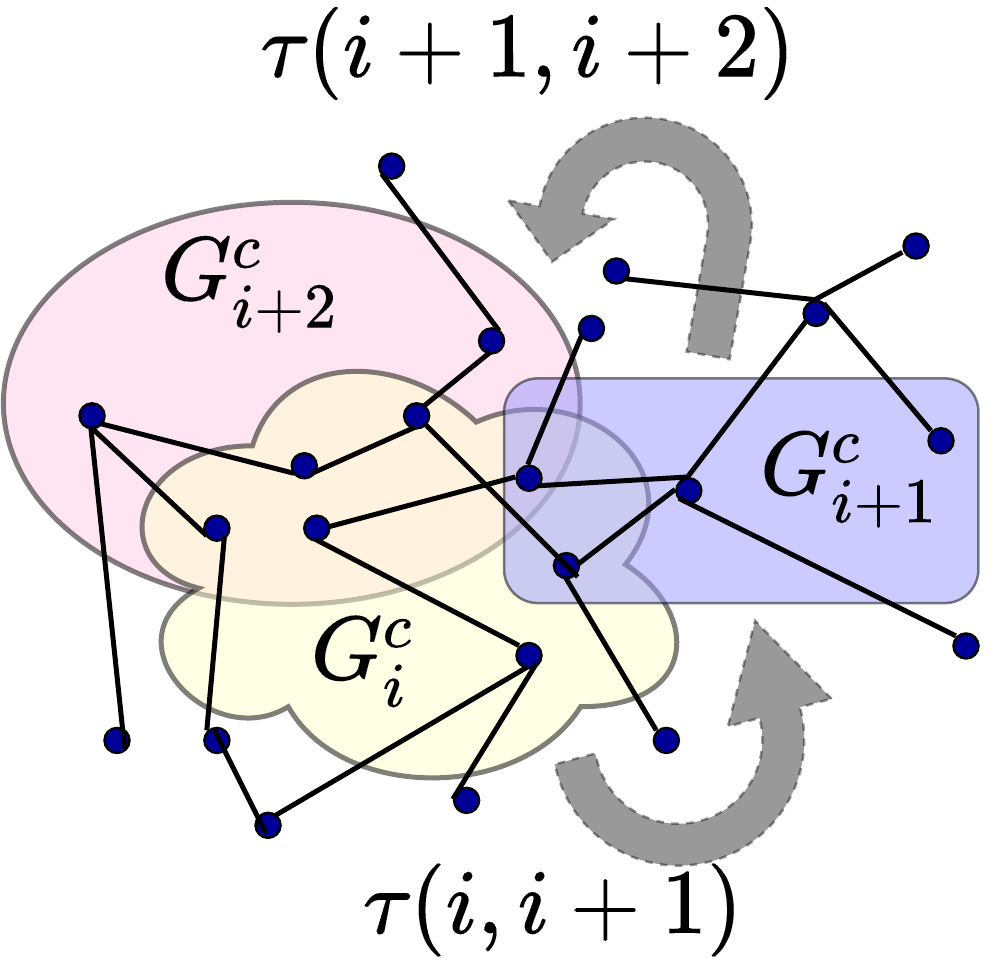}} \label{subFig:c1}
}~~~~~~~~~~
\subfigure[Case C: Situation-2]{
{\includegraphics[trim=278 416 38 96, clip,width=0.20\textwidth]{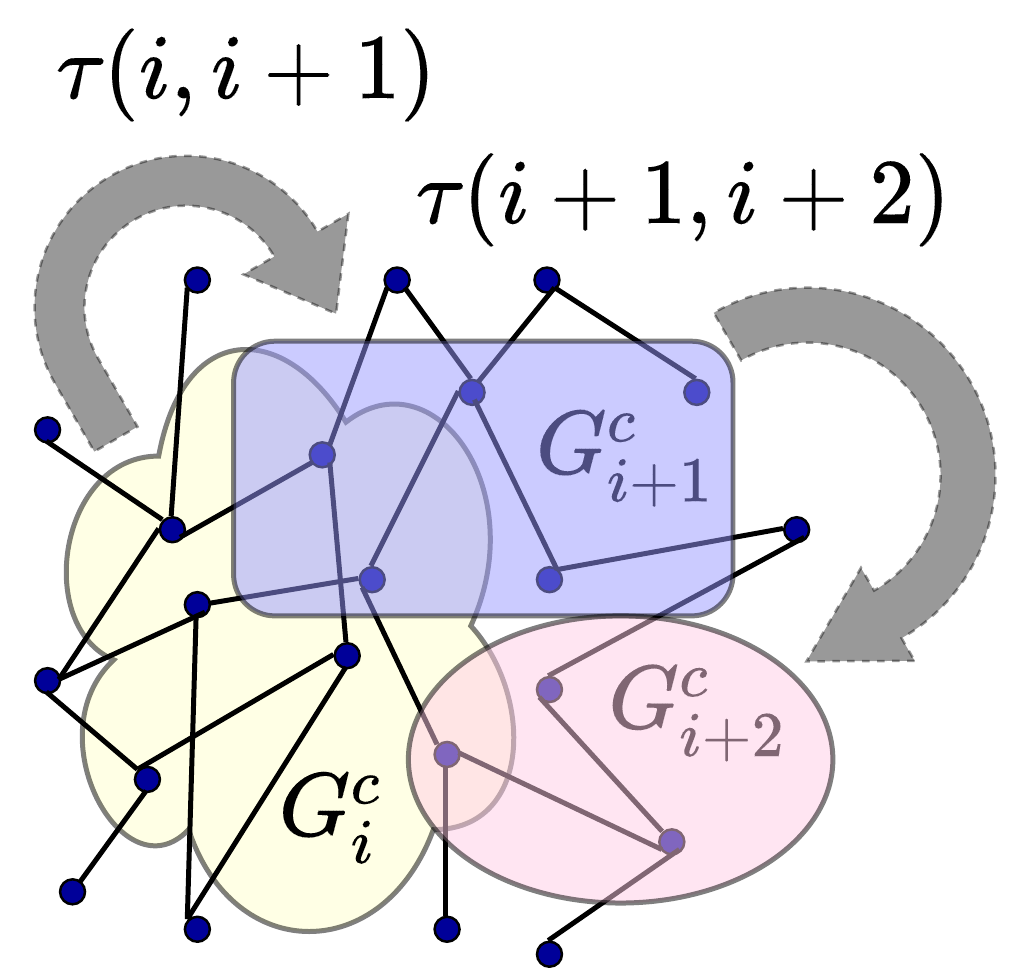}} \label{subFig:c2}
}\\
\subfigure[Case C: Situation-3]{
{\includegraphics[trim=195 372 130 127, clip,width=0.20\textwidth]{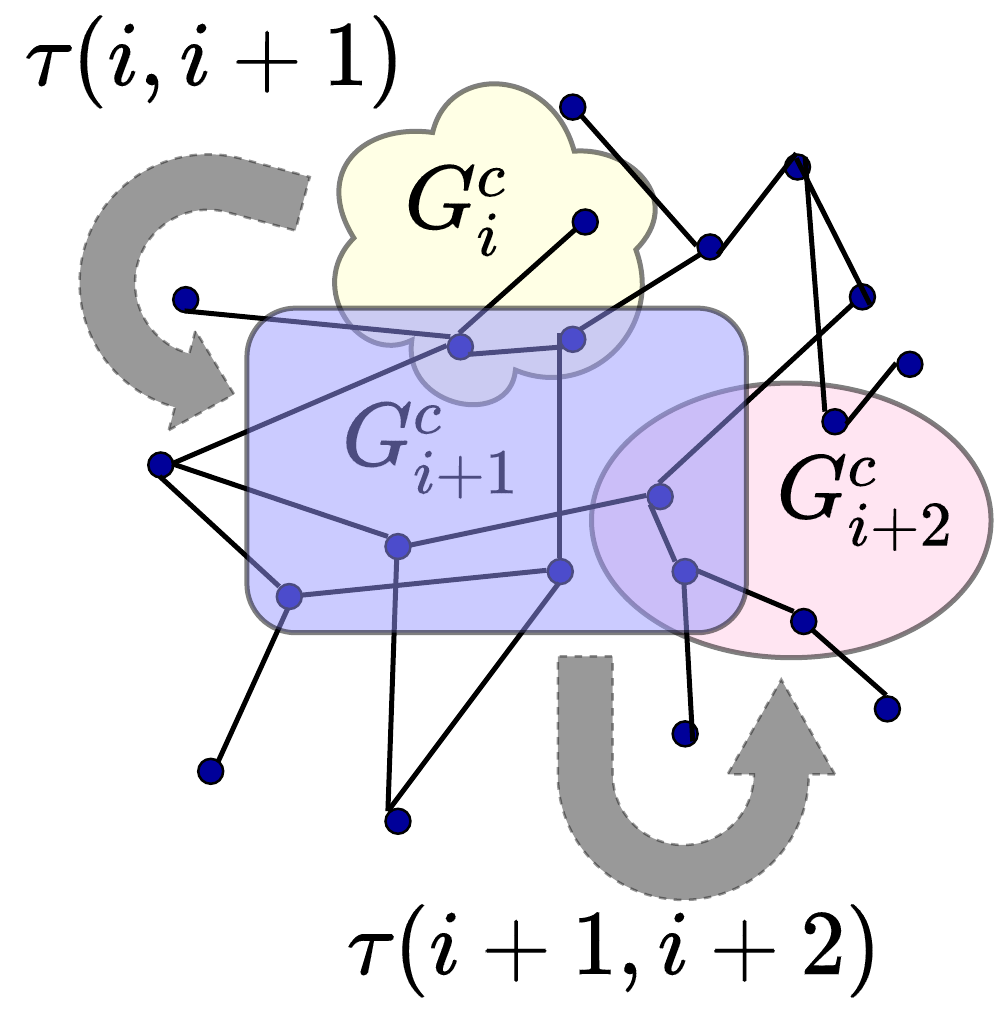}} \label{subFig:c3}
}~~~~~~~~~~
\subfigure[Case C: Situation-4]{
{\includegraphics[trim=232 410 100 97, clip,width=0.20\textwidth]{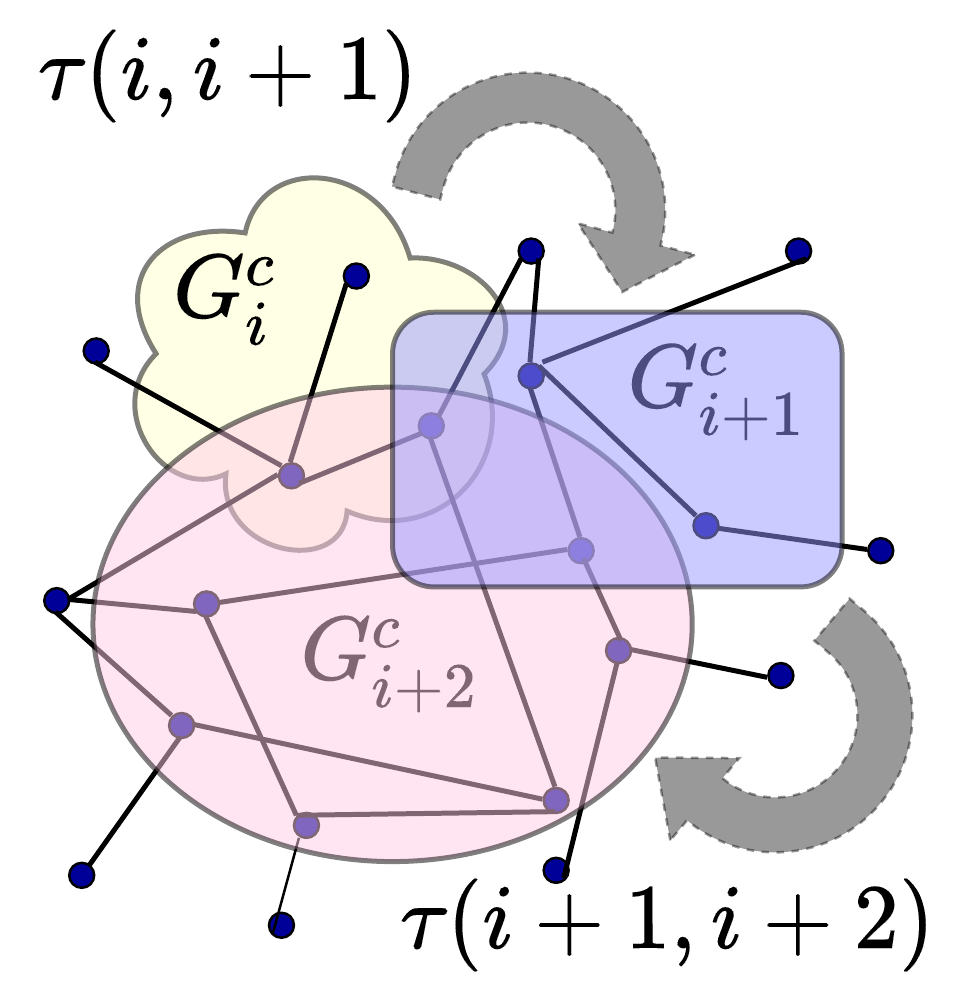}} \label{subFig:c4}
}
\caption{An  illustration of the scenario (Case C), occurs during teaching.} \label{fig:caseC}
\end{figure}

\end{enumerate}

To handle these situations, we measure how different concepts are densely connected to each other within a topic. For a graph $G$, we define graph density metric $\delta(G)$ as, $\delta(G) =  \frac{\sum_{k=1}^{N} w(e_{k})}{N(N-1)}$, while, $w(e_{k})$ is the weight of edge $e_{k}$ and $N$ is the total number of vertices in $G$. We take different combinations of \textit{cluster\_centroid}, compute their density and compare these density values with each other to find the actual topic boundaries. Here the idea is, concepts belong in same topic have more contextual semantic similarity than they are in different topics. Combined graph, denoted as, $\mathcal{C}(G^{c}_{i},G^{c}_{i+1}) = (V',E')$ is formed by combining $G^{c}_{i} = (V^{c}_{i},E^{c}_{i})$ and $G^{c}_{i+1} = (V^{c}_{i+1},E^{c}_{i+1})$ where, $V' = \{V^{c}_{i} \cup V^{c}_{i+1}\}$ and $E' = \{E^{c}_{i} \cup E^{c}_{i+1} \cup \mathcal{E}\}$. Here, $\mathcal{E}$ represents set of additional edges present in the $\mathscr{KG}$ between any pair of concepts that belongs in $G^{c}_{i}$ or $G^{c}_{i+1}$. Using these density measure we compute actual topic boundaries using the algorithm shown in Algorithm~\ref{algo:tpBoundary}.

\begin{algorithm*}[!ht]
\caption{Compute Topic Boundary List} \label{algo:tpBoundary}
\begin{algorithmic}[1]
\State $tpBoundaries \leftarrow [~], start\_time \leftarrow centroidList[0].startTime, end\_time \leftarrow 0, i \leftarrow 0 $ 
\State $len \leftarrow centroidList.length$
\While{$i < len-2$ }
  \State $c0 \leftarrow centroidList[i], ~c1 \leftarrow centroidList[i+1], ~c2 \leftarrow centroidList[i+2]$ 
  
  \If{$c0.vertices \cap c1.vertices = \phi ~\And~ c1.vertices \cap c2.vertices = \phi$} \textcolor{blue}{\Comment Case A}
    \State $end\_time \leftarrow c0.endTime$
    \State $tpBoundaries \leftarrow tpBoundaries \cup (start\_time,end\_time)$ \textcolor{blue}{\Comment ${\tau(i,i+1)}$ is topic boundary}
    \State $start\_time \leftarrow c1.startTime$ \textcolor{blue}{\Comment set start\_time value for the next topic}
    \State $i \leftarrow i+1$
    \State $continue$
  \EndIf
  \newline  \textcolor{blue}{\Comment compute density of different combination of \textit{cluster\_centroid} as described}
  
  \State $m01 \leftarrow combine(c0,c1), ~delta01 = density(m01)$
  \State $m02 \leftarrow combine(c0,c2), ~delta02 = density(m02)$
  \State $m12 \leftarrow combine(c1,c2), ~delta12 = density(m12)$
  \State $m123 \leftarrow combine(c1,c2,c3), ~delta123 = density(m123)$

  \If{$delta02 > delta01 ~\And~ delta02 > delta12$} \textcolor{blue}{\Comment Case C: Situation-1}
    \State $i \leftarrow i+2$
    \State $continue$
  
  \ElsIf{$delta123 > delta01 ~\Or~ delta123 > delta12 ~\Or~ delta123 > delta02$} \textcolor{blue}{\Comment Case C: Situation-2,3,4}
    \State $i \leftarrow i+2$
    \State $continue$
  
  \ElsIf{$delta01 > delta12$} \textcolor{blue}{\Comment Case B: Situation-1}
    \State $end\_time \leftarrow c1.endTime$
    \State $tpBoundaries \leftarrow tpBoundaries \cup (start\_time,end\_time)$ \textcolor{blue}{\Comment ${\tau(i+1,i+2)}$ is topic boundary}
    \State $start\_time = c2.startTime$ \textcolor{blue}{\Comment set start\_time value for the next topic}
    \State $i \leftarrow i+2$
    \State $continue$
  
  \ElsIf{$delta12 > delta01$} \textcolor{blue}{\Comment Case B: Situation-2}
    
    \State $end\_time \leftarrow c0.endTime$
    \State $tpBoundaries \leftarrow tpBoundaries \cup (start\_time,end\_time)$ \textcolor{blue}{\Comment ${\tau(i,i+1)}$ is topic boundary}
    \State $start\_time \leftarrow c1.startTime$ \textcolor{blue}{\Comment set start\_time value for the next topic}
    \State $i \leftarrow i+1$
    \State $continue$
  \EndIf
  
\EndWhile
\State $end\_time \leftarrow centroidList[len-1].endTime$ \textcolor{blue}{\Comment taking end-time of last cluster}
\State $tpBoundaries \leftarrow tpBoundaries \cup (start\_time,end\_time)$
\Return $tpBoundaries$
\end{algorithmic}
\end{algorithm*}

 Lines 5-10 handle \textit{Case A} by computing vertex intersection between three consecutive \textit{cluster\_centroid} and mark ${\tau(i,i+1)}$ as topic boundary. \textit{Case B} and \textit{Case C} are determined by comparing density of different combined graph formed by $G^{c}_{i}$, $G^{c}_{i+1}$ and $G^{c}_{i+2}$. For \textit{Case B:Situation-1}, graph density of $\mathcal{C}(G^{c}_{i},G^{c}_{i+1})$ is more than density of $\mathcal{C}(G^{c}_{i+1},G^{c}_{i+2})$ as $G^{c}_{i}$ and $G^{c}_{i+1}$ both belong in same topic and $G^{c}_{i+1}$ and $G^{c}_{i+2}$ belong in two different topics. Hence ${\tau(i+1,i+2)}$ is a topic boundary. These are handled in lines 21-26. In similar manner  \textit{Case B:Situation-2} is handled in lines 27-32. In \textit{Case C:Situation-1}, concepts covered in $G^{c}_{i}$ and $G^{c}_{i+2}$ are almost similar but there is some concept change in $G^{c}_{i+1}$. So $\mathcal{C}(G^{c}_{i},G^{c}_{i+2})$ will have higher density value than that of $\mathcal{C}(G^{c}_{i+1},G^{c}_{i+2})$ and $\mathcal{C}(G^{c}_{i},G^{c}_{i+1})$ which is handled in lines 15-17. For \textit{Case C:Situation-2,3,4} all $G^{c}_{i}$,$G^{c}_{i+1}$ and $G^{c}_{i+2}$ are in same topic, so density of $\mathcal{C}(G^{c}_{i},G^{c}_{i+1},G^{c}_{i+2})$ is higher than the individual density of $\mathcal{C}(G^{c}_{i},G^{c}_{i+1})$ or $\mathcal{C}(G^{c}_{i},G^{c}_{i+2})$ or $\mathcal{C}(G^{c}_{i+1},G^{c}_{i+2})$. These kind of situations are handled in lines 18-20.

\section{Annotation} \label{sec:annotation}
At this stage we assign topic name to each video segment obtained from previous stage. We describe step by step procedure to accomplish this task using both visual and textual semantic resources, the video file and the topic name list.

\subsubsection{Text Line Identification and Recognition}
\begin{figure}[!h]

\centering
\subfigure[]{
    \includegraphics[trim=65 45 50 45, clip,width=0.21\textwidth]{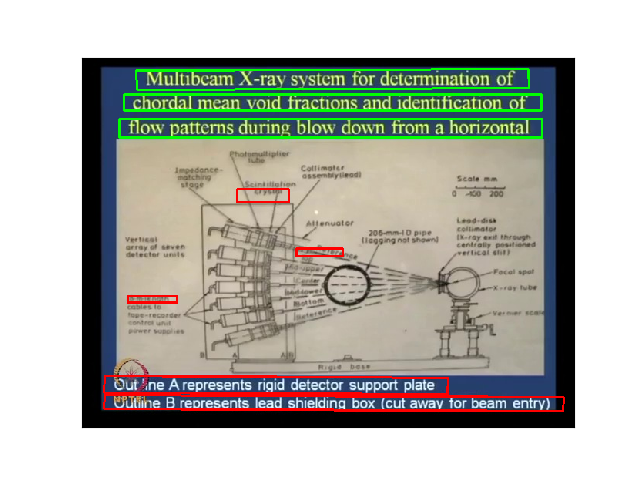}
    \label{fig:text_line_ex_b}
}~
\subfigure[]{
    \includegraphics[trim=65 45 50 45, clip,width=0.21\textwidth]{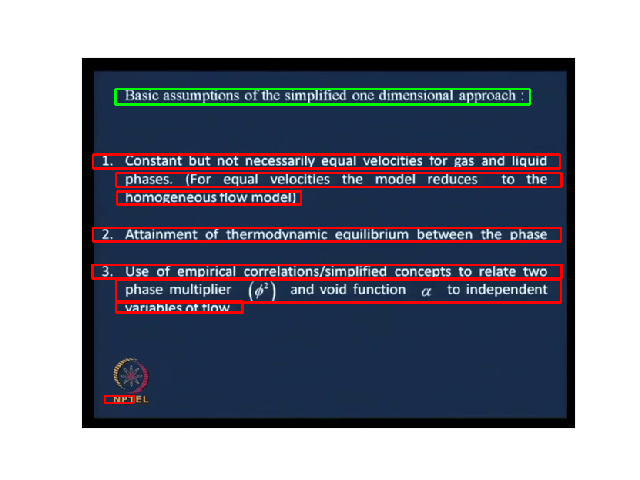}
    \label{fig:text_line_ex_c}
}
\subfigure[]{
    \includegraphics[trim=65 45 50 45, clip,width=0.21\textwidth]{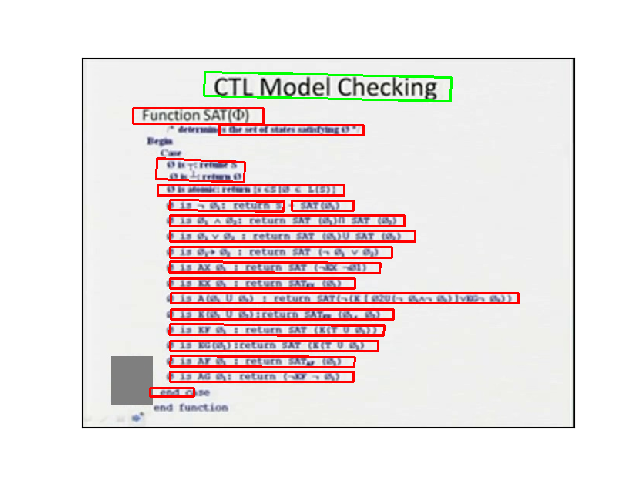}
    \label{fig:text_line_ex_d}
}~
\subfigure[]{
    \includegraphics[trim=65 45 50 45, clip,width=0.21\textwidth]{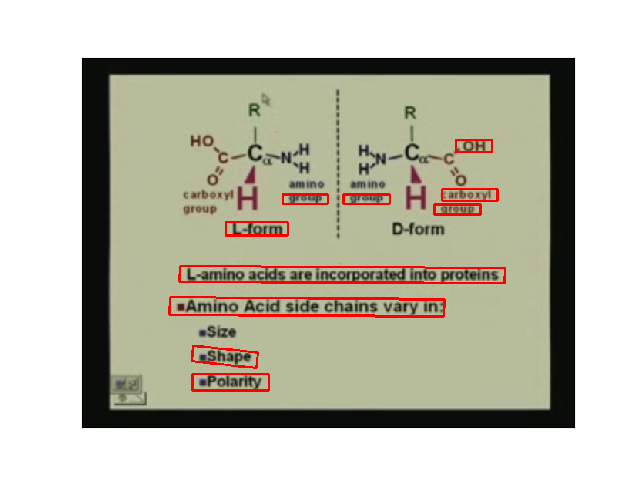}
    \label{fig:text_line_ex_c}
}

\caption{Some sample slides with identified text line with  detecting title (green box) and non-title (red box)}

\label{fig:text_line_ex}

\end{figure}

In this step we extract text line from each unique slide frames. Text Line identification is done using Stroke Width Transform (SWT) \cite{epshtein2010detecting}. SWT is an local image operator that computes the width of the most likely stroke of a pixel and this operation is done for all the pixel. The output of SWT is an image equal to the input image size, each element contains the width of the stroke associated with the pixel. Then, by leveraging some natural observations on text line like, it appears in linear form, similarities in stroke width of the characters, letter width, height and spaces between the letters and words, SWT creates a bounding box surrounding each text line. Next, cropped text line images are passed to an OCR engine for text recognition. Open source OCR library \textit{tessaract} \cite{tess} is used for it. Some slide images with detected text lines are shown in figure \ref{fig:text_line_ex}.
 

\subsubsection{Determine Topic Name}
\begin{figure}[!h]

\centering

\includegraphics[width=0.45\textwidth]{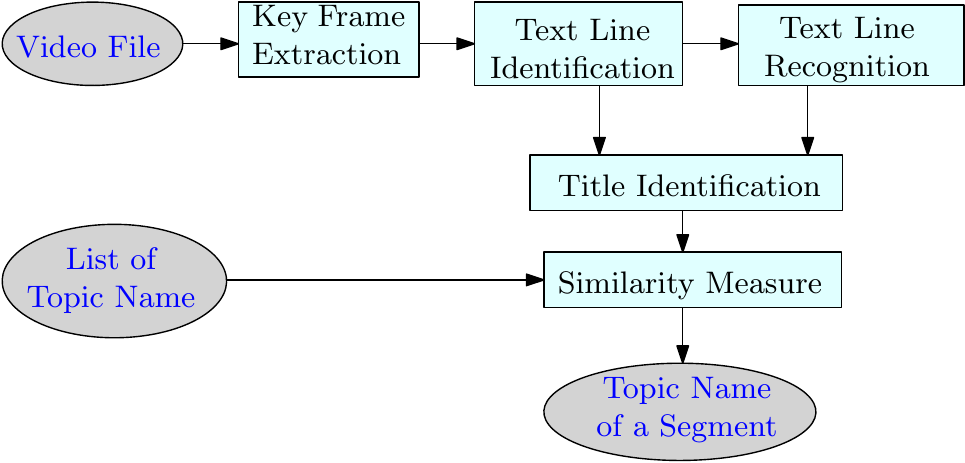}

\caption{Schematic diagram of the annotation process}

\label{fig:Annotation}

\end{figure}

For a video segment, this step finds proper topic name using the course syllabus and identified slide titles. First we remove duplicate titles, punctuation marks and apply lemmatization on each slide title. Let us assume, within time stamp $TS_{p}$ and $TS_{q}$ we get $N$ unique titles denoted as $t_{1}$, $t_{2}$ ... $t_{N}$. Now for each topic name, we measure cumulative similarity score between that topic name and all $N$ titles. Topic name that have highest cumulative similarity score is assigned to this video segment. Once a topic name is assigned, we remove that name from the topic name list so that same name is not assigned to multiple video segment. Word Mover's Distance \cite{kusner2015word} is used for similarity measure. Using these methods we can successfully annotate each video segment obtained previously. Schematic diagram of annotation process is shown in Figure \ref{fig:Annotation}

\section{Fusion} \label{sec:fusion}

In this step we have two set of segmentation result obtained from two different approach mentioned above \textit{i.e.} using structural analysis of the transcript file an another is from analyzing the knowledge semantics of the transcribed text leveraging domain knowledge graph. We found that structural analysis base segmentation approach performs well for small length topic segments while for large length segments this approach misses the segment boundary by large margin that reduces the overall accuracy of the topic segmentation. The situation becomes even worse for the lecture videos which mostly have comparative large topic segments. Experimental results of this shown in Section~\ref{subsec:eval}. To alleviate this problem we use segmentation list obtained from both the approach to create the final segmentation result. Our result merging approach is given as follows.

\begin{enumerate}
 \item For small length topic segments, we take average starting time and ending time from both the segmentation results to make the final results.
 \item For large length topic segments, we discard the segmentation information obtained from the syntactic structure based result and only consider the results obtained from the semantic analysis.
\end{enumerate}

Now, we need to determine the threshold value for deciding small sized topic segments and large sized topics. We take this value as $15$ minutes. We select this value experimentally and the related experiments are shown in the Section~\ref{subsec:threshold}. 

\section{Experiments}

\subsection{Dataset ans Ground Truth}
In this work we use NPTEL lecture videos. These videos are freely available in NPTEL website. The dataset has the following properties. 

\begin{enumerate}
 \item NPTEL offers different courses, each contains $\sim$40 one hour long lecture videos.
 \item Each lecture video contains $\sim$40 one hour long lecture videos. Each video is synchronized with lecture slides, \textit{i.e.} at the time of lecture, instructor uses power-point/beamer presentation which is displayed in the video.
 \item Each video is associated with speech-to-text transcript file. NPTEL also maintains course syllabus containing different topics taught in the whole course.
 \item Usually each video consists of multiple topics taught sequentially and each topic constitutes of multiple concepts as chosen by the instructor.
 \item For evaluate the structural segmentation approach we select $10$ different subjects from engineering discipline and take $10$ lecture videos from each of them.  The subject name and the corresponding topic counts are shown in the Table ~\ref{tab:10Subjs}.
 
 \begin{table}[!ht]

\caption{Sample Ground Truth} \label{tab:10Subjs}
\centering
\resizebox{0.46\textwidth}{!}{

\begin{tabular}{|l|l|}
\hline
\textbf{Subject Name}                                         & \textbf{\begin{tabular}[c]{@{}l@{}}Topic \\ Count\end{tabular}} \\ \hline
Digital Image Processing                             & 74                                                        \\ \hline
Data Structures and Algorithms                       & 61                                                        \\ \hline
Design Verification and test of Digital VLSI Designs & 54                                                        \\ \hline
Principles of Physical Metallurgy                    & 79                                                        \\ \hline
Cryptography and Network Security                    & 57                                                        \\ \hline
Low Power VLSI Circuits and Systems                  & 57                                                        \\ \hline
Building Materials and Construction                  & 88                                                        \\ \hline
Optimal Control Guidance and Estimation              & 62                                                        \\ \hline
Advanced Control System Design                       & 72                                                        \\ \hline
Industrial Engineering                               & 65                                                        \\ \hline
\end{tabular}
}

\end{table}
 \item For evaluating semantic segmentation we take $65$ lecture videos from $2$ courses from software engineering having $255$ topics as shown in Table~\ref{tab:semanticSubjs}.
 
 \begin{table}[!ht]

\caption{Sample Ground Truth} \label{tab:semanticSubjs}
\centering
\resizebox{0.31\textwidth}{!}{
\begin{tabular}{|l|l|}
\hline
\textbf{Subject Name}                                                               & \textbf{\begin{tabular}[c]{@{}l@{}}Topic \\ Count\end{tabular}} \\ \hline
\begin{tabular}[c]{@{}l@{}}Fundamentals of\\ embedded software testing\end{tabular} & 102                                                             \\ \hline
Software Engineering                                                                & 155                                                             \\ \hline
\end{tabular}
}

\end{table}
 
\end{enumerate}

Ground truth are prepared by the course instructor and/or teaching assistants of the corresponding courses.  Ground truth is prepared by consulting the lecture videos and the course syllabus available in NPTEL website. Ground truth of a lecture video consists of several topics along with their start-time and end-time. An overview of the ground truth is shown in Table~\ref{tab:sampleGT}.

\begin{table}[!ht]

\caption{Sample Ground Truth} \label{tab:sampleGT}
\centering
\resizebox{0.46\textwidth}{!}{
\begin{tabular}{lcc}
\hline
\textbf{Topic Name} & \textbf{\begin{tabular}[c]{@{}l@{}}Start time\\ (mm : ss)\end{tabular}} & \textbf{\begin{tabular}[c]{@{}l@{}}End time\\ (mm : ss)\end{tabular}}  \\ \hline
Requirement Analysis          & 00:32                                                                 & 15:43                                                               \\ 
Design         & 15:46                                                                 & 27:12                                                               \\ 
Implementation      & 15:46                                                                 & 37:50                                                               \\ 
Verification           & 38:00                                                                 & 54:33                                                               \\ 
Maintenance & 54:37   & 59:43 \\
\hline
\end{tabular}
}

\end{table}

\subsection{Evaluation} \label{subsec:eval}
We determine the topic name of a video segment by matching the textual similarities between slide titles and topic list present in course syllabus. Slide titles are extracted using standard methods described by Yang et al. \cite{yang2014content}. Then, we measure cumulative similarity score between these titles and all topic names present in course syllabus. Topic name that have highest similarity score is assigned to the video segment under consideration. Once a topic name is assigned, we remove that name from the topic name list so that same name is not assigned to multiple video segments. We use Word Mover's Distance \cite{kusner2015word} for the similarity measure.

For evaluation, we measure the similarity between topic intervals generated from our system and intervals present in the ground truth. To find interval similarity of a topic, we measure the overlapping time ratio (\textit{otr}) between ground truth and system generated interval information for a given topic. Conceptually \textit{otr} is similar as Jaccard Similarity \cite{jaccard1908nouvelles} and we define it as, $$ \textit{otr} =  \frac{\min (eTime_{gr}, eTime_{sys}) - \max(sTime_{gr}, sTime_{sys})}{\max(eTime_{gr}, eTime_{sys}) - \min(sTime_{gr}, sTime_{sys})} $$ Where, ($sTime_{gr}$,  $eTime_{gr}$) is the time interval of a topic in ground truth and ($sTime_{sys}$, $eTime_{sys}$) is the time interval generated by the system. For multiple topics, we take arithmetic mean of \textit{otr} for all the topics. We perform holistic evaluation of all the lecture videos in our dataset and measure similarity score of them. Table~\ref{tab:titleIdentComp} shows similarity score of different lecture series. Also for the evaluation we use standard segmentation measurement metric like $P_{k}$ \cite{beeferman1997text} and WindowDiff \cite{pevzner2002critique}.

\begin{table*}[!h]

\centering
\caption{Comparison using F1 score and OTR}
\label{tab:titleIdentComp}
\resizebox{0.89\textwidth}{!}{
\begin{tabular}{|c|l|c|c|c|c|}
\hline
\multirow{2}{*}{\textbf{\begin{tabular}[c]{@{}c@{}}Subj \\ ID\end{tabular}}} & \multicolumn{1}{c|}{\multirow{2}{*}{\textbf{Subject Name (Number of Topics)}}} & \multicolumn{2}{c|}{\textbf{F1 Score}}                                                                       & \multicolumn{2}{c|}{\textbf{Overlapping Time Ratio}}                                                         \\ \cline{3-6} 
                                                                             & \multicolumn{1}{c|}{}                                                          & \textbf{\begin{tabular}[c]{@{}c@{}}Syntactic Structure\\  Analysis\end{tabular}} & \textbf{Semantic Analysis} & \textbf{\begin{tabular}[c]{@{}c@{}}Syntactic Structure\\  Analysis\end{tabular}} & \textbf{Semantic Analysis} \\ \hline
1                                                                            & Software Testing (123)                                                          & 0.72                                                                             & 0.78                      & 0.75                                                                             & 0.82                      \\ \hline
2                                                                            & Software Engineering (75)                                                      & 0.71                                                                             & 0.76                      & 0.73                                                                             & 0.81                      \\ \hline
3                                                                            & Software Project Management (68)                                               & 0.75                                                                             & 0.79                      & 0.79                                                                             & 0.84                      \\ \hline
\multicolumn{1}{|l|}{}                                                       & \textbf{ALL}                                                                   & 0.73                                                                             & 0.78                      & 0.76                                                                             & 0.83                      \\ \hline
\end{tabular}
}
\end{table*}

\begin{table*}[!h]

\centering
\caption{Comparison using Pk and WindowDiff}
\label{tab:titleIdentComp_2}
\resizebox{0.89\textwidth}{!}{
\begin{tabular}{|c|l|c|c|c|c|}
\hline
\multirow{2}{*}{\textbf{\begin{tabular}[c]{@{}c@{}}subj\\ ID\end{tabular}}} & \multicolumn{1}{c|}{\multirow{2}{*}{\textbf{Subject Name}}} & \multicolumn{2}{c|}{\textbf{Pk}}                                                                             & \multicolumn{2}{c|}{\textbf{WinDiff}}                                                                        \\ \cline{3-6} 
                                                                            & \multicolumn{1}{c|}{}                                       & \textbf{\begin{tabular}[c]{@{}c@{}}Syntactic Structure\\  Analysis\end{tabular}} & \textbf{Semantic Analysis} & \textbf{\begin{tabular}[c]{@{}c@{}}Syntactic Structure\\  Analysis\end{tabular}} & \textbf{Semantic Analysis} \\ \hline
1                                                                           & Software Testing (123)                                       & 0.37                                                                             & 0.32                      & 0.31                                                                             & 0.26                      \\ \hline
2                                                                           & Software Engineering (75)                                   & 0.31                                                                             & 0.29                      & 0.28                                                                             & 0.22                      \\ \hline
3                                                                           & Software Project Management (68)                            & 0.34                                                                             & 0.31                      & 0.33                                                                             & 0.28                      \\ \hline
\multicolumn{1}{|l|}{}                                                      & \textbf{ALL}                                                & 0.34                                                                             & 0.31                      & 0.31                                                                             & 0.25                      \\ \hline
\end{tabular}
}
\end{table*}

\begin{table*}[!h]
\centering
\caption{Segmentation accuracy comparison of different sized topic segments}
\label{tab:comp_contri}
\resizebox{0.89\textwidth}{!}{

\begin{tabular}{|l|c|c|c|c|c|c|}
\hline

\multicolumn{1}{|c|}{\multirow{2}{*}{\textbf{Subject Name}}} & \multicolumn{3}{c|}{\textbf{Topic Duration \textless{}= 15 minutes}}            & \multicolumn{3}{c|}{\textbf{Topic Duration \textgreater 15 minutes}}            \\ \cline{2-7} 

\multicolumn{1}{|c|}{}                                       & \textbf{No of topic} & \textbf{syntactic analysis} & \textbf{semantic analysis} & \textbf{No of topic} & \textbf{syntactic analysis} & \textbf{semantic analysis} \\ \hline

Fundamentals of embedded software testing                                    & 52                   & 0.79                        & 0.82                       & 50                   & 0.57                        & 0.84                       \\ \hline

Software Engineering                                         & 83                   & 0.83                        & 0.78                       & 59                   & 0.49                        & 0.82                       \\ \hline

ALL                                                          & 135                  & 0.82                        & 0.81                       & 109                  & 0.54                       & 0.83                       \\ \hline

\end{tabular}
}
\end{table*}

\subsection{Threshold Value selection} \label{subsec:threshold}

We use bi-directional LSTM with attention layer in the structural segmentation method. Now as LSTM have some limitations of capturing the context of the text corpora, here in the transcript file it captures the underlined sentence meaning in local context and not able to get the global context. So with this shortcomings our model is not able to accurately detect the segment boundaries of large duration topics. As mentioned earlier in Section~\ref{sec:semantic} we use semantic segmentation based approach using knowledge graph to obtain more accurate topic boundary for topics with duration greater than a certain threshold level. Here we we make some experiments to determine these threshold values of the topic duration. 

\begin{figure}[!h]
\centering
\includegraphics[width=0.45\textwidth]{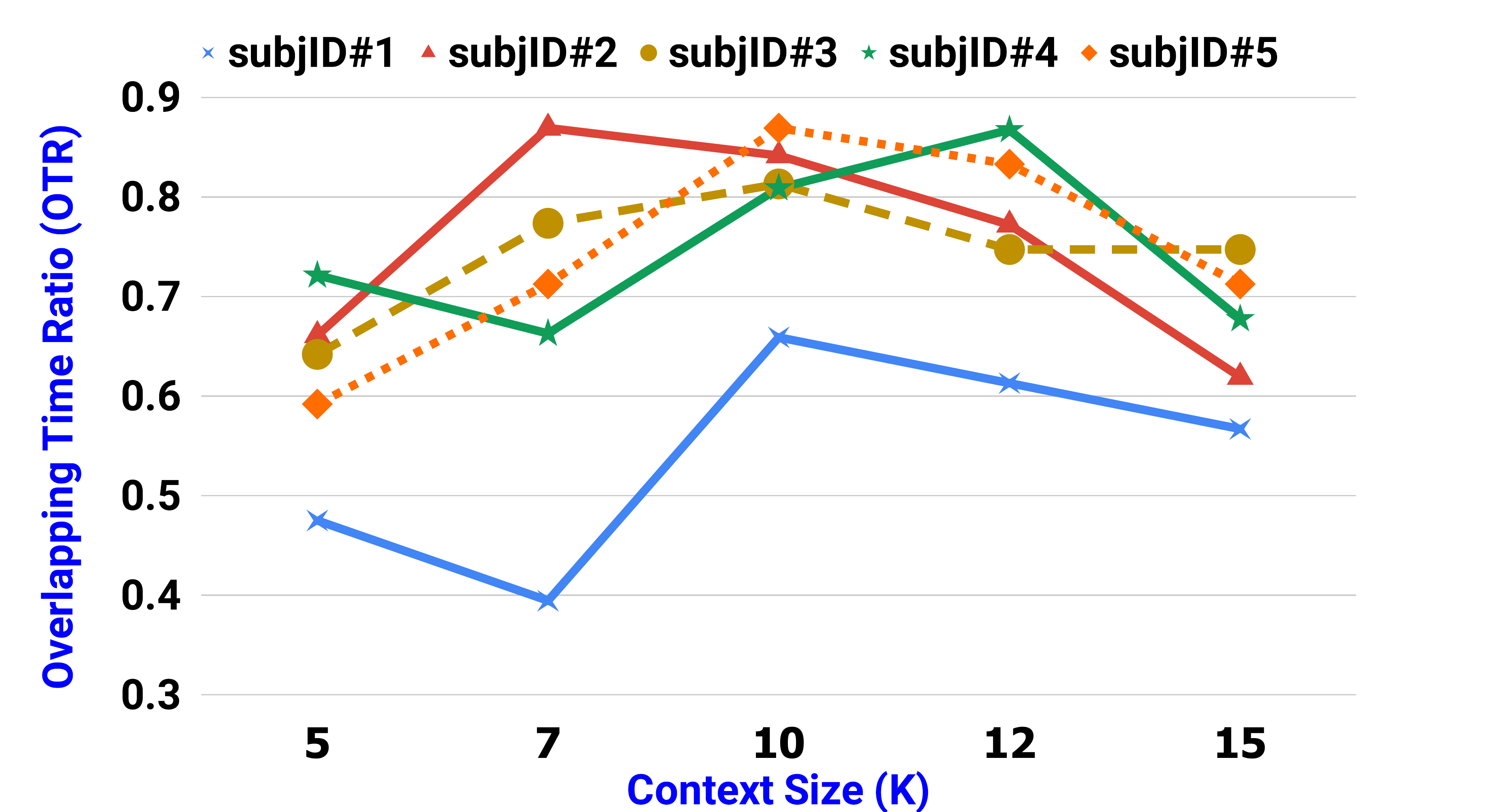}
\caption{Context Size vs OTR Score}
\label{fig:K-chart}
\end{figure}

First we determine the optimal context size used in the sentence classifier described in Section~\ref{subsec:sentClass}. For finding the optimal context size we make a holistic evaluation of the segmented topics and compare the results with the ground truth. The context size $K$ should be such that the neighbourhood information is sufficient enough to predict accurate segment boundary as well as takes minimal time to train the model. Higher value of $K$ captures some unnecessary context while too lowering this value may not properly capture the original context. We optimize the value of $K$ by measuring the segmentation accuracy using OTR metric on the dataset. Figure~ \ref{fig:K-chart} shows the experimental result on different values of $K$. After experiment we take the value of $K$ (number of sentences in each context) as $10$ with $20$ words in each sentence. 

To determine the threshold value of topic duration, we perform some experiments that will make some comparative analysis between structural segmentation and semantic segmentation based on topic duration. From the $66$ lecture videos on software engineering domain we get $225$ topics. We divide these topics into two category according to their duration. We find $164$ topics has duration less than $15$ minutes while $91$ topics have duration more than $15$ minutes. We measure the segmentation accuracy on these two group of topics and find the results as shown in the Table~\ref{tab:comp_contri}.

The reason of this discrepancy between two approach for long duration topic is, structural analysis primarily focuses on the local context on a text block, more specifically it produce best result for particular context size. If the lecture video duration is greater than the context size, the underlined model is not able to capture the global context. If the topic duration is smaller than the context size the model ignores some important context. Use of knowledge graph and semantic analysis alleviates this problem and perform better for the large length topic segments. For more illustration, we have divided these $255$ topics obtained from $65$ videos into four different category based on their duration. the duration categories with the corresponding number of topics are given below. \newline

123 topic with duration 10 minutes to 15 minutes. \newline

123 topic with duration 15 minutes to 20 minutes. \newline

123 topic with duration 20 minutes to 25 minutes. \newline

123 topic with duration more than 25 minutes. \newline

\begin{figure}[!h]
\centering
\caption{Comparison between structural analysis and semantic analysis on large length topics.}

 \label{fig:comp_ana}

 {\includegraphics[trim=0 0 120 0, clip, width=0.45\textwidth]{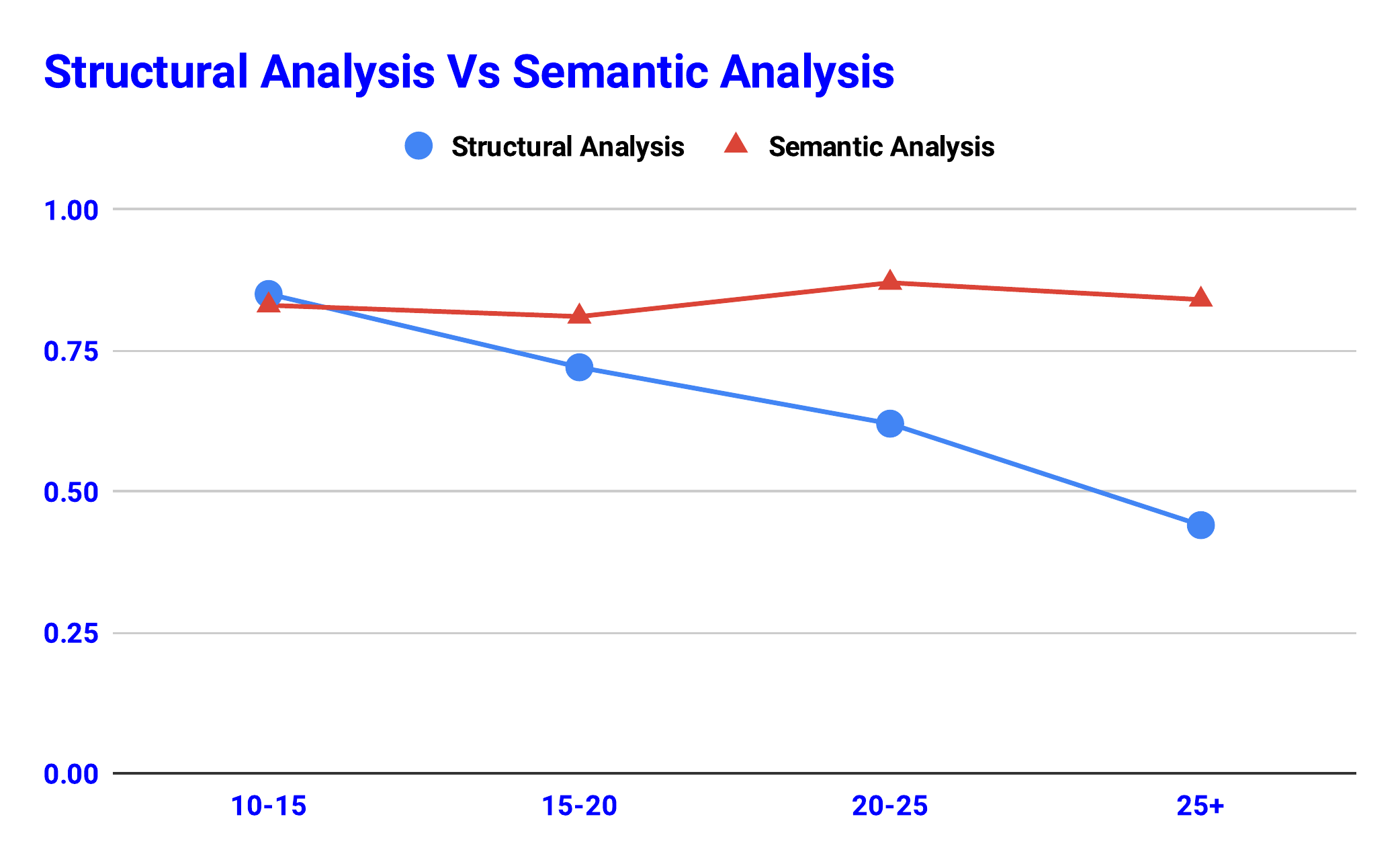}}
\end{figure}

We measure group wise segmentation accuracy of two different algorithms  proposed  by us and Figure~\ref{fig:comp_ana} shows the comparative results. It is clear from the figure that as the segment duration increases accuracy of structural analysis based method decreases. While on the other hand, semantic analysis based approach produces steady performance irrespective of the topic size.

\section{Conclusion}

In this work we detect the topic boundaries of long MOOC lecture videos. These topics can be used for indexing purpose on which user can perform topic search that reduce user time of locating and browsing a topic inside a lecture video. For doing this we use a state-of-the art language model to capture boundary information. Also we leverage the inherent knowledge present inside a doamin knowledge graph for fine tuning the segmentation result and accurately detecting the topics of larger time duration. In this work we have used LSTM with attention layer as the language model and BERT for capturing the overall semantics of the text corpora. In future we have a plan to use transformaer instead of LSTM and use domain specific BERT model for getting better segmentation result.


%





\ifCLASSOPTIONcaptionsoff
  \newpage
\fi



\bibliographystyle{IEEEtran}
\bibliography{refs_ananda}
%

%

\end{document}